\documentclass[11pt]{article}

\usepackage[preprint]{acl}

\usepackage{times}
\usepackage{latexsym}
\usepackage{enumitem}
\usepackage[T1]{fontenc}
\usepackage[utf8]{inputenc}

\usepackage{microtype}
\usepackage{inconsolata}
\usepackage{microtype}
\usepackage{inconsolata}
\usepackage{graphicx}
\usepackage{amsmath,amssymb,mathtools}

\usepackage{booktabs,tabularx,array,multirow}
\usepackage{algorithm}
\usepackage{algpseudocode}
\usepackage{placeins}
\usepackage{xcolor}
\usepackage{colortbl}
\usepackage{float}

\usepackage{graphicx}

\newcommand{\R}{\mathbb{R}}
\newcommand{\softmax}{\operatorname{softmax}}
\newcommand{\Mean}{\operatorname{Mean}}

\newcommand{\na}{\textnormal{n/a}}
\newcommand{\method}{TILA-CT}
\newcolumntype{Y}{>{\raggedright\arraybackslash}X}
\definecolor{tilablue}{RGB}{31,86,124}
\definecolor{tilalight}{RGB}{234,242,248}
\definecolor{tilagreen}{RGB}{232,243,235}
\definecolor{tiladelta}{RGB}{234,244,252}
\definecolor{sdgray}{gray}{0.35}

\title{Self-Indexing Latent Memories for Access-Aware \\ Continual Instruction Tuning of LLMs}

\author{
Dhruv Dixit\textsuperscript{2*}
Paritosh Pandey\textsuperscript{1*}
Snehalkumar `Neil' S. Gaikwad\textsuperscript{1,2*} \\
\\
\textsuperscript{*}Society-Centered AI Lab \\
\textsuperscript{1}Department of Computer Science \\
\textsuperscript{2}School of Data and Information Sciences \\
University of North Carolina, Chapel Hill, USA \\
dhruvd@unc.edu \quad
pandeyp@cs.unc.edu \quad
gaikwad@cs.unc.edu
}

\begin{document}

\maketitle

\begin{abstract}
Continual instruction tuning often preserves old skills with task-private modules. Inference still requires selecting the right module from the prompt alone. We introduce Task-Isolated Latent Adapter Continual Training (\method), a replay-free method that makes each task module self-indexing. Each task receives a latent memory and adapter bank in a frozen LLM. During adaptation, the memory gathers causal-prefix information and injects task-conditioned context. For access, its persistent state provides the routing prototype used by a shared prompt-only router. On the 8-task CT-Pipeline stream, \method\ achieves 65.19 acquisition and 63.98 final macro score, with 1.38 points of average forgetting. A middle-reversed order gives 65.38 acquisition and 64.17 final macro score. Across 6 backbones, final scores remain within 1.5 points of acquisition. Ablations support memory-derived routing keys and causal gathering. Final macro routing accuracy is 93.9\%, and oracle routing recovers 1.21 final macro points. These results separate preserved task computation from learned access to it.
\end{abstract}
\section{Introduction}

In continual instruction tuning, a language model receives supervised instruction-output tasks one stage at a time. In the task-incremental setting, task boundaries are known during training, each stage uses only current-task data, and inference does not provide task identities. The model must retain earlier task adaptations and choose the correct preserved computation path from the instruction and input.

Shared parameter updates can overwrite earlier adaptations, causing catastrophic forgetting \citep{mccloskey1989catastrophic,french1999catastrophic}; recent LLM continual-learning studies report the same issue in instruction tuning streams \citep{wang2023trace,zhang2023citb}. Task-private modules reduce this interference by freezing completed parameters. Their performance still depends on access. A frozen task path is useful only when the router selects it for the right input. We therefore separate \emph{task-path preservation} from \emph{module access}: preservation fixes the computation available for a task, while access determines whether inference uses it.

Existing task-incremental continual instruction tuning methods address this setting with replay, regularization, prompts, adapters, or expert expansion. Modular PEFT methods are closest to our work. MoCL composes frozen PEFT modules through input-to-task matching \citep{wang2024mocl}. SAPT connects adaptation and selection through shared attention \citep{zhao2024sapt}. SwitchCIT learns a task classifier \citep{wu2024switchcit}. GoD-MoE expands and reuses routed LoRA experts to control capacity growth \citep{zhang2026growondemand}. These methods use modular adaptation with learned inference-time selection. We study one access question inside this family: how should the state used for task adaptation identify the module to its router?

We propose Task-Isolated Latent Adapter Continual Training (\method), which makes task modules \emph{self-indexing}. Each task receives a private latent memory and an adapter bank in the upper layers of a frozen language model. During task-conditioned computation, the memory gathers information from each token's causal prefix and injects task-specific context through the adapters. The same persistent memory is mean-pooled to form the routing prototype for that task. Task isolation provides preservation, causal latent gather-and-inject provides adaptation, and memory-derived routing provides access. At inference, a shared prompt-only router selects one observed module for the entire sequence.

The core design contribution is the use of one learned state for both adapter computation and module identification. Task isolation, learned module selection, latent attention, and task-conditioned adapters have all appeared in prior work (Section~\ref{sec:related}). After each task, the complete private path is frozen. Later training updates the shared routing projections and adds new candidate keys, so old-task predictions can change through module selection even when the corresponding task path is preserved. We measure this behavior using routing accuracy, routing degradation, and the gap to task-oracle selection.

On CT-Pipeline, \method\ reaches 65.19 acquisition and 63.98 final macro performance, with 1.38 points of average forgetting. It leads 6 of 8 final tasks despite 5 acquisition deficits. The ScienceQA deficit falls from 4.32 points at acquisition to 0.48 points at final evaluation. Across 6 backbones, final scores remain within 1.5 points of acquisition. A middle-reversed order gives similar retention for \method\ and the 2 strongest final-performance baselines. These results support an access-aware analysis of task learning and retention.

Our contributions are:
\begin{itemize}[itemsep=1mm, parsep=0pt]
    \item We introduce self-indexing latent memories for replay-free continual instruction tuning of frozen LLMs. The same task-private memory conditions causal adapter computation and supplies its module’s routing prototype, enabling module selection without task labels at inference.
    \item We test the design with ablations on memory-derived keys, causal gathering, latent adaptation, residual scaling, and task isolation.
    \item We analyze acquisition, final performance, routing degradation, oracle routing, storage growth, task-order sensitivity, and cross-backbone behavior.
\end{itemize}
\section{Related Work}
\label{sec:related}

\paragraph{Continual instruction tuning for LLMs.}
Prior surveys organize continual learning for LLMs around continued pre-training, domain adaptation, and fine-tuning settings \citep{shi2025continual}. We place \method\ under task-incremental continual instruction tuning, where tasks arrive sequentially and inference proceeds without task labels. Prior CIT methods reduce forgetting through replay, regularization, prompt expansion, adapter isolation, or expert modules. In modular CIT, final performance also depends on assigning an input to the correct frozen task path. We study this access problem with self-indexing task memories.

\paragraph{Continual adaptation and isolation.}
EWC, Synaptic Intelligence, and Learning without Forgetting constrain shared parameters or outputs \citep{kirkpatrick2017overcoming,zenke2017continual,li2018learning}. GEM, A-GEM, and GPM constrain updates using stored examples, gradients, or feature subspaces \citep{lopezpaz2017gradient,chaudhry2019efficient,saha2021gradient}. NLP studies address generated replay, instruction streams, and knowledge adaptation \citep{sun2020lamol,ke2022continual,zhang2023citb,seo2024train}. Progressive Neural Networks and PackNet use architectural isolation \citep{rusu2016progressive,mallya2018packnet}. We use isolation for preservation and measure router access to frozen task paths.

\paragraph{Modular PEFT and task access.}
Prompts, adapters, and low-rank updates reduce trainable parameters \citep{lester2021power,houlsby2019parameter,hu2022lora,dettmers2023qlora}. AdapterFusion combines adapters \citep{pfeiffer2021adapterfusion}, and L2P retrieves learned prompts \citep{wang2022learning}. MoCL freezes completed PEFT modules and composes their outputs through input-to-task matching without test-time task identities \citep{wang2024mocl}. SAPT jointly learns task modules and an attention-based selector \citep{zhao2024sapt}. SwitchCIT selects private adapters with a task classifier \citep{wu2024switchcit}. These methods motivate access-aware modular tuning. We derive the routing prototype from the same private latent memory used inside the adapter computation. The module's adaptation state also identifies it to the router.

\begin{figure*}[t]
  \centering
  \includegraphics[width=0.90\linewidth]{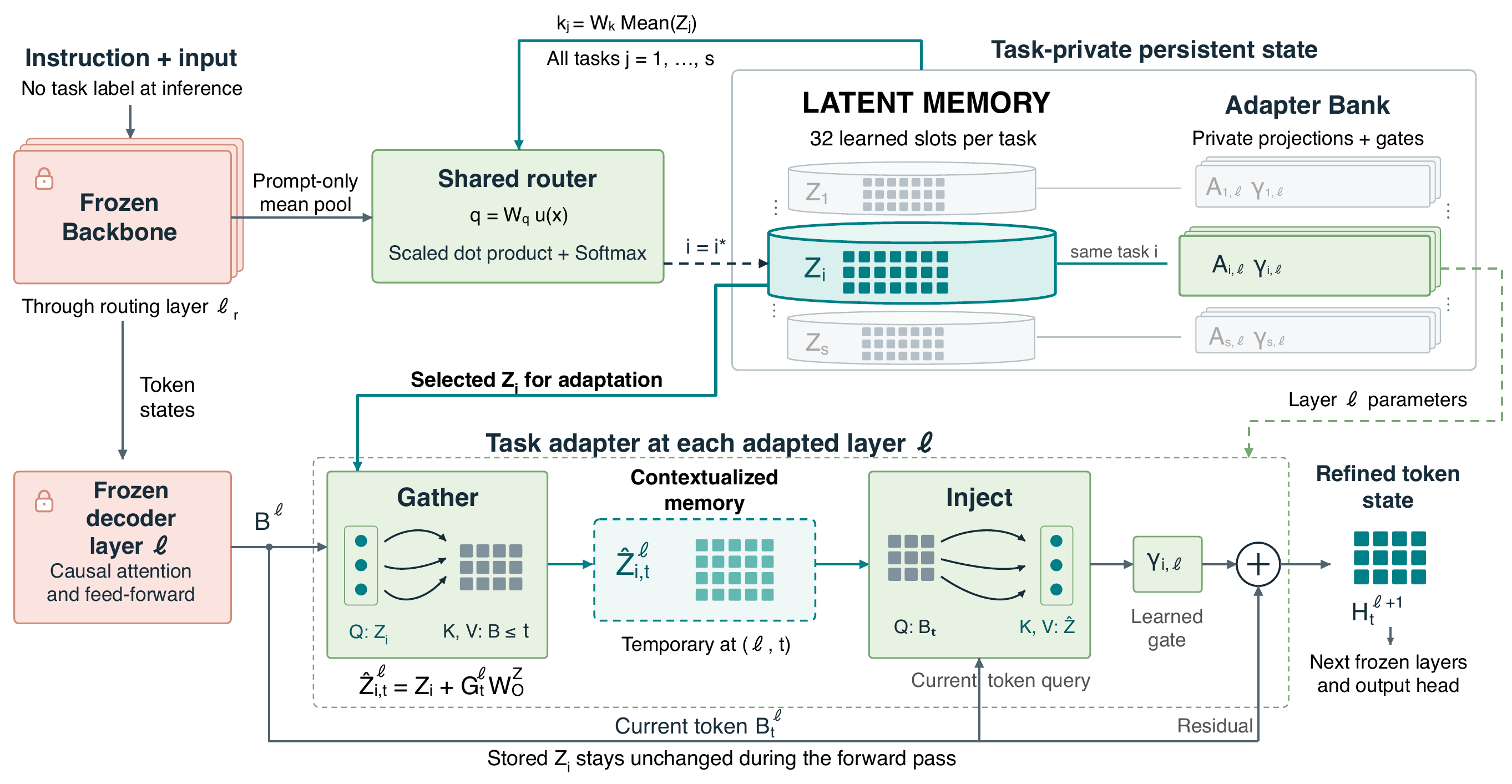}
\caption{Overview of TILA-CT. A frozen backbone produces prompt features for routing and token states for adaptation. The shared router selects one task-private module using a key derived from the task's persistent latent memory. The selected memory gathers causal-prefix information and injects task-conditioned context through adapters at each adapted layer. Pink marks frozen backbone computation, green marks learned router and adapter computation, and gray marks stored modules or intermediate representations.}
  \label{fig:TILA-CT-architecture}
\end{figure*}

\paragraph{Expansion and interference control.}
GoD-MoE uses replay-free routed LoRA experts with selective expansion, freezing, and reuse \citep{zhang2026growondemand}. TreeLoRA organizes reusable adaptation state through gradient similarity and hierarchical search \citep{qian2025treelora}. InfLoRA constrains updates with protected feature subspaces \citep{liang2024inflora}. GainLoRA adds input-dependent gates and constrains new gates to limit their effect on earlier inputs \citep{liang2025gainlora}. These methods study capacity growth, expert reuse, or functional interference. We use one complete private module per task and measure how accurately the router selects that module. Among replay-free modular CIT methods, \method\ is closest to GoD-MoE. GoD-MoE studies expert reuse and expansion. \method\ studies memory-derived access to frozen task paths.

\paragraph{Latent attention and task conditioning.}
Set Transformer's inducing-point attention gathers information into learned slots and returns it to input tokens \citep{lee2019settransformer}. Perceiver IO uses latent read and write operations \citep{jaegle2022perceiverio}. Flamingo combines latent resampling with gated injection into a frozen language model \citep{alayrac2022flamingo}. HyperFormer generates adapters from task embeddings \citep{mahabadi2021hyperformer}. These works use learned latent state to condition model computation. We use causal gather-and-inject attention inside task-private continual adapters and use the same persistent latent array as the routing key source. The ablations compare this construction with independent keys, standard adapters, and static-memory injection.

\section{Access-Aware Continual Instruction Tuning}
\label{sec:method}

\subsection{Setting and persistent state}

We consider task-incremental continual instruction tuning with known training-task boundaries. Tasks $\mathcal D_1,\ldots,\mathcal D_K$ arrive sequentially. At stage $s$, training uses only examples from $\mathcal D_s$. No stored or generated examples from earlier tasks are used. At inference, the model receives an instruction and input without a task identity and must select among the $s$ observed modules.

The backbone $f_{\theta_0}$, including its token embeddings and output head, remains frozen. Each task $i$ receives a private state
\begin{equation}
\mathcal P_i=\bigl(Z_i,\{A_{i,\ell},\gamma_{i,\ell}\}_{\ell\in\mathcal L}\bigr),
\label{eq:private-state}
\end{equation}
where $Z_i\in\R^{L_z\times d}$ is a learned latent memory and $\mathcal L$ is the set of adapted decoder layers. For Qwen3-8B-Reasoning, $\mathcal L$ contains 7 upper decoder layers. The adapters $A_{i,\ell}$ and residual gates $\gamma_{i,\ell}$ are private to task $i$. Stage $s$ trains the current memory, current adapters and gates, and shared routing projections $W_q,W_k$. Completed task modules remain frozen.

\subsection{Routing from adapter memories}

The router reads frozen hidden states at layer $\ell_r$, immediately before the first adapted layer. These states do not depend on the selected module. Let $\rho_t(x)$ equal 1 for an instruction or input token and 0 for a response or padding token. The prompt summary is
\begin{equation}
u(x)=\frac{\sum_t\rho_t(x)H_t^{\ell_r}}{\sum_t\rho_t(x)}\in\R^d.
\label{eq:router-pooling}
\end{equation}
During training, the known response boundary defines the mask. During inference, the mask includes every non-padding prompt token.

Each module is indexed by the pooled persistent memory $\Mean(Z_i)$. This memory also conditions the adapter computation below. The routing prototype differs from the temporary prefix-conditioned memory used during gather-and-inject computation. After task $i$, $Z_i$ is frozen, while its projected key can still change through updates to the shared projection $W_k$.

The router maps the prompt summary to a query and each task memory to a key:
\begin{align}
q_s(x)&=W_q^{(s)}u(x),\quad
k_i^{(s)}=W_k^{(s)}\Mean(Z_i),
\label{eq:keys}\\
\alpha_i^{(s)}(x)&=\softmax_i\left(
\frac{q_s(x)^\top k_i^{(s)}}{\sqrt{d_r}}
\right),
\label{eq:router}\\
i_s^*(x)&=\arg\max_{1\leq i\leq s}\alpha_i^{(s)}(x).
\end{align}
At inference, the router compares all available keys and selects one module for the whole sequence. Inference uses no task label and no reference response. During training, the known task identity provides the routing label. Updates to $W_q$ and $W_k$, together with newly added keys, can change access to earlier modules.

\subsection{Gather, inject, and residual fusion}

Let $B^\ell\in\R^{T\times d}$ be the output of the frozen attention and feed-forward transformations at adapted decoder layer $\ell$. At token position $t$, the adapter reads only the causal prefix $B^\ell_{\leq t}$. Both cross-attention operations use 8 heads of width $d_h=48$, giving total adapter width $d_a=384$. We use standard scaled dot-product attention \citep{vaswani2017attention}:
\begin{equation}
\operatorname{Attn}(Q,K,V)=
\softmax\!\left(\frac{QK^\top}{\sqrt{d_h}}\right)V.
\label{eq:head-attention}
\end{equation}

Below, $\operatorname{Attn}$ concatenates the head outputs, and the task-layer indices on private projections are omitted. The gather step uses the persistent memory as queries and the causal prefix as keys and values:
\begin{align}
G_t^\ell
&=\operatorname{Attn}_{\mathrm g}
\left(
Z_iW_Q^Z,
B^\ell_{\leq t}W_K^H,
B^\ell_{\leq t}W_V^H
\right),\\
\widehat Z_{i,t}^\ell
&=Z_i+G_t^\ell W_O^Z.
\label{eq:gather}
\end{align}
The temporary memory $\widehat Z_{i,t}^\ell$ is specific to layer $\ell$ and token position $t$. The stored memory $Z_i$ is unchanged during the forward pass.

The inject step uses the current token as a query over this temporary memory:
\begin{align}
J_t^\ell
&=\operatorname{Attn}_{\mathrm i}
\left(
B_t^\ell W_Q^H,
\widehat Z_{i,t}^\ell W_K^Z,
\widehat Z_{i,t}^\ell W_V^Z
\right),\\
H_t^{\ell+1}
&=B_t^\ell+\gamma_{i,\ell}J_t^\ell W_O^H.
\label{eq:inject}
\end{align}

The stored memory is shared across adapted layers. Each adapted layer has private attention projections. Each residual gate starts at 0.1 and is learned during task training. Appendix~\ref{app:method-details} gives tensor shapes and token flow.

\paragraph{Causal visibility.}
Training and inference use the same prefix restriction. Under shifted causal-LM supervision, the state at position $t$ predicts token $t+1$. The gather step at $t$ can read token states only through $t$. During generation, it reads the prompt and the tokens generated so far. The response-only loss mask determines which next-token predictions receive supervision.

\subsection{Training and conditional preservation}

Task identity is used only as a routing label during training. It is excluded from the model input. The objective combines response-token language modeling with task-selection supervision:
\begin{equation}
\mathcal L_s
=
\mathcal L_{\mathrm{LM}}(\mathcal D_s)
-\lambda_{\mathrm r}\log\alpha_s(x),
\qquad
\lambda_{\mathrm r}=0.1.
\label{eq:objective}
\end{equation}
The language-model loss supervises response targets only. The routing loss trains $W_q$, $W_k$, and the current memory from prompt-only features. Hard argmax is used for module selection at inference and has no gradient through the selected index. During stage $s$, the trainable parameters are restricted to the current private module and the shared router.

For a fixed task index $j$, the task-conditioned computation depends on the frozen backbone and completed module $\mathcal P_j$. Freezing these parameters preserves the output distribution for the same prefix and input processing. Shared router updates can still change which module is selected. Oracle routing forces the correct module and measures the preserved path. Learned routing measures access to that path from the prompt.

\section{Experimental Design}
\label{sec:experiments}

\paragraph{Datasets.}
Our primary CT-Pipeline order is FOMC (monetary-policy stance classification)~\citep{shah2023trillion}; C-STANCE (Chinese stance classification)~\citep{zhao2023c}; MeetingBank (meeting summarization)~\citep{hu2023meetingbank}; ScienceQA (science question answering)~\citep{lu2022learn}; NumGLUE-en (numerical reasoning)~\citep{mishra2022numglue}; JEC-QA (legal QA)~\citep{zhong2020jec}; MedMCQA (medical QA)~\citep{pal2022medmcqa}; and 20Minuten (German text simplification)~\citep{kew202320}. Tasks overlap with TRACE~\citep{wang2023trace} but follow a different order. Examples contain instruction, input, and output fields. Metrics are accuracy or normalized exact match, except ROUGE-L F1 for MeetingBank and SARI for 20Minuten. These heterogeneous metrics use a 0 to 100 scale and receive equal macro weight. Appendix~\ref{app:data} gives splits and response extraction.

\paragraph{Backbones.}
We evaluate \method\ on six frozen backbones spanning model scales and dense Transformer, sparse mixture-of-experts, and hybrid Transformer--Mamba2 architectures: Llama-3.2-1B~\citep{meta2024llama32model}, Qwen-1.8B~\citep{qwen2023qwen18bmodel}, Llama-3.1-8B~\citep{meta2024llama31model}, OLMoE-7B~\citep{allenai2024olmoemodel}, Qwen3-8B~\citep{qwen2025qwen3model} (\texttt{enable\_thinking=True}), and Falcon-H1R-7B~\citep{tii2026falconh1rmodel}. Table~\ref{tab:backbones} reports acquisition and final retention.

\paragraph{Protocol.}
On the main backbone, Qwen3-8B-Reasoning, each task receives one epoch of response-token supervision with Adafactor at $2\times10^{-5}$, effective batch size four, and optimizer state resets between stages. Maximum training/evaluation lengths are 256/512 tokens. \method\ uses seven adapted layers, 32 memory slots, eight adapter heads of width 48 (total 384), and routing width 256 (Appendix~\ref{app:implementation}). Each checkpoint is evaluated on all observed tasks; framework task scores average seeds 37, 73, and 101. We also rerun \method, GoD-MoE, and GainLoRA with stages 2--7 reversed, endpoints fixed, and the same protocol; other baselines use the primary order (Appendix~\ref{app:protocol}). Capacity sweeps vary latent memory size $L_z$ and adapter attention width $d_a$, reporting acquisition/final macro scores and final task scores for selected configurations.

\paragraph{Baselines.}
We compare \method\ with sequential fine-tuning (Seq-Train), L2P~\citep{wang2022learning}, GEM~\citep{lopezpaz2017gradient}, A-GEM~\citep{chaudhry2019efficient}, SAPT~\citep{zhao2024sapt}, InfLoRA~\citep{liang2024inflora}, QLoRA-CT, SwitchCIT~\citep{wu2024switchcit}, MoCL~\citep{wang2024mocl}, TreeLoRA~\citep{qian2025treelora}, GainLoRA~\citep{liang2025gainlora}, and GoD-MoE~\citep{zhang2026growondemand}. The last four are replay-free and require no inference task labels. GEM/A-GEM retain 256 examples per task; SAPT uses generated inputs for attention regularization without old-task language-model loss; SwitchCIT retains examples for classifier training; QLoRA-CT selects adapters by reference-output loss. All share the one-epoch task budget and use comparable reported adaptation capacities where supported; auxiliary procedures, total and active computation, information access, and total stored state differ (Appendix~\ref{app:protocol}). Table~\ref{tab:qwen-config} lists layer, rank, and target-module configurations.

\begin{table*}[t]
\centering
\small
\setlength{\tabcolsep}{3pt}
\renewcommand{\arraystretch}{1.08}
\resizebox{0.75\textwidth}{!}{%
\begin{tabular}{@{}lrrrrrrc@{}}
\toprule
Method & $D$ & $P$ & $\overline F$ & $\operatorname{Var}(F)$ & Mean $|\Delta|$ & Gain [95\% CI] & Wins \\
\midrule
Seq-Train & $51.37_{\color{sdgray}\pm 0.13}$ & $45.43_{\color{sdgray}\pm 0.11}$ & 6.79 & 22.46 & 7.42 & 18.55 [14.17, 22.14] & 8/8 \\
L2P       & $50.43_{\color{sdgray}\pm 0.12}$ & $45.27_{\color{sdgray}\pm 0.10}$ & 5.89 & 18.78 & 4.41 & 18.71 [15.72, 21.90] & 8/8 \\
GEM       & $48.87_{\color{sdgray}\pm 0.12}$ & $44.52_{\color{sdgray}\pm 0.11}$ & 4.97 & 15.39 & 5.21 & 19.46 [14.65, 24.95] & 8/8 \\
A-GEM     & $50.61_{\color{sdgray}\pm 0.14}$ & $46.66_{\color{sdgray}\pm 0.12}$ & 4.51 & 11.15 & 4.27 & 17.32 [12.35, 23.20] & 8/8 \\
SAPT      & $64.19_{\color{sdgray}\pm 0.16}$ & $60.81_{\color{sdgray}\pm 0.15}$ & 3.87 & 5.88  & 1.19 & 3.17 [0.61, 6.24] & 6/8 \\
InfLoRA   & $63.20_{\color{sdgray}\pm 0.14}$ & $59.73_{\color{sdgray}\pm 0.13}$ & 3.97 & 5.61  & 1.76 & 4.25 [2.51, 6.32] & 8/8 \\
SwitchCIT & $64.11_{\color{sdgray}\pm 0.15}$ & $60.41_{\color{sdgray}\pm 0.14}$ & 4.24 & 7.71  & 1.35 & 3.57 [1.47, 5.94] & 7/8 \\
QLoRA-CT  & $59.43_{\color{sdgray}\pm 0.13}$ & $55.26_{\color{sdgray}\pm 0.15}$ & 4.76 & 10.51 & 2.52 & 8.72 [6.49, 11.06] & 8/8 \\
MoCL      & $60.84_{\color{sdgray}\pm 0.48}$ & $57.05_{\color{sdgray}\pm 0.13}$ & 4.33 & 7.54  & 2.09 & 6.93 [5.20, 8.76] & 8/8 \\
TreeLoRA  & $62.25_{\color{sdgray}\pm 0.53}$ & $58.84_{\color{sdgray}\pm 0.16}$ & 3.89 & 5.07  & 1.64 & 5.14 [3.90, 6.46] & 8/8 \\
GainLoRA  & $63.58_{\color{sdgray}\pm 0.41}$ & $60.70_{\color{sdgray}\pm 0.14}$ & 3.29 & 5.05  & 1.36 & 3.28 [2.53, 4.15] & 8/8 \\
GainLoRA \textit{(mid-rev.)} & $63.46_{\color{sdgray}\pm 0.41}$ & $60.89_{\color{sdgray}\pm 0.14}$ & 2.94 & 5.06 & 1.35 & 3.28 [2.47, 4.05] & 8/8 \\
GoD-MoE   & $63.65_{\color{sdgray}\pm 0.41}$ & $62.01_{\color{sdgray}\pm 0.17}$ & 1.88 & 2.06  & 1.21 & 1.97 [1.41, 2.58] & 8/8 \\
GoD-MoE \textit{(mid-rev.)}  & $63.81_{\color{sdgray}\pm 0.41}$ & $61.89_{\color{sdgray}\pm 0.17}$ & 2.19 & 2.05 & 1.21 & 2.28 [1.43, 2.58] & 8/8 \\
\midrule
\textbf{\method} & $\mathbf{65.19}_{\color{sdgray}\pm 0.15}$ & $\mathbf{63.98}_{\color{sdgray}\pm 0.17}$ & \textbf{1.38} & \textbf{0.76} & \textbf{0.72} & -- & -- \\
\textit{\method\ (mid-rev.)} & $\mathbf{65.38}_{\color{sdgray}\pm 0.15}$ & $\mathbf{64.17}_{\color{sdgray}\pm 0.17}$ & \textbf{1.39} & \textbf{0.76} & \textbf{0.72} & -- & -- \\
\bottomrule
\end{tabular}%
}
\caption{Results on Qwen3 backbone (mean over seeds 37, 73, 101). Gray subscripts are standard deviations. Forgetting and its variance use 7 mean task losses. Mean $|\Delta|$ excludes task 1. \textit{(mid-rev.)} means that order was used.}
\label{tab:mainresults}
\end{table*}

\paragraph{Metrics.}
Let $R_{s,j}$ be the three-seed mean score on task $j$ after stage $s$. For $K=8$ tasks, acquisition $D$, final performance $P$, and forgetting satisfy
\begin{gather}
D=\frac1K\sum_jR_{j,j},\qquad P=\frac1K\sum_jR_{K,j},\\
F_j=R_{j,j}-R_{K,j},\qquad\overline F=\frac1{K-1}\sum_{j<K}F_j,
\label{eq:forgetting}\\
P=D-\frac{K-1}{K}\overline F.\label{eq:decomposition}
\end{gather}
Negative $F_j$ indicates improvement after acquisition; $\operatorname{Var}(F)$ is the population variance of the first seven $F_j$. Signed acquisition gaps $\Delta_j=R_{j,j}-S_j^{\mathrm{iso}}$ use isolated scores $S_j^{\mathrm{iso}}$, obtained from each method's isolated variant for framework comparisons. Mean $|\Delta|$ excludes task 1 and measures reference proximity. All summaries and margins, including variance, maximum loss, and absolute gaps, use seed-averaged task scores before rounding. For each primary-order baseline, we bootstrap eight paired final-score differences (\method{} minus baseline), drawing eight with replacement 10,000 times (seed 42). Table~\ref{tab:mainresults} reports mean gains, 95\% percentile CIs, and wins (strictly positive differences) out of eight. Appendix~\ref{app:reference-details} pairs SEQ, ISO, and their differences.

\paragraph{Ablations.}
Qwen3-8B-Reasoning ablations comprise A0 (full \method), A1 (oracle routing), A2 (independent routing keys), A3 (parameter-matched standard adapters), A4 (injection without gathering), A5 (fixed residual scaling), A6 (shared adapter bank), and A7 (single shared module). A1 reuses A0 checkpoints without retraining; A2/A3 share independent keys for a controlled latent-versus-standard adapter comparison (Appendix~\ref{app:ablation-details}).

\section{Results}

\subsection{Strong acquisition with low subsequent loss}
\label{sec:mainresults}

\begin{table*}[t]
\centering\small\setlength{\tabcolsep}{6pt}
\begin{tabular}{@{}lrrrrrrr@{}}
\toprule
& \multicolumn{3}{c}{Acquisition and reference proximity} & \multicolumn{4}{c}{Final retention}\\
\cmidrule(lr){2-4}\cmidrule(l){5-8}
Backbone & $D$ & Mean $|\Delta|$ & Max $|\Delta|$ & $P$ & $\overline F$ & BWT & $\operatorname{Var}(F)$\\
\midrule
Llama-3.2-1B        & 67.07 & 1.09 & 1.51 & 65.83 & 1.42 & $-1.42$ & 0.78 \\
Qwen-1.8B           & 65.67 & 1.09 & 1.92 & 64.47 & 1.37 & $-1.37$ & 0.77 \\
Llama-3.1-8B        & 69.14 & 1.12 & 1.56 & 67.86 & 1.46 & $-1.46$ & 0.81 \\
OLMoE-7B            & 70.12 & 1.86 & 2.93 & 68.86 & 1.44 & $-1.44$ & 0.82 \\
Qwen3-8B-Reasoning  & 65.19 & 0.72 & 1.31 & 63.98 & 1.38 & $-1.38$ & 0.76 \\
Falcon-H1R-7B       & 77.88 & 0.82 & 0.90 & 76.42 & 1.67 & $-1.67$ & 0.91 \\
\bottomrule
\end{tabular}
\caption{Cross-backbone acquisition and final retention. Reference-gap summaries exclude task 1. Appendix~\ref{app:reference-details} provides paired SEQ and ISO scores. Average forgetting and BWT for the additional backbones are derived from the reported $D$ and $P$ using Equation~\ref{eq:decomposition}; their task-level variances are reported summaries.}
\label{tab:backbones}
\end{table*}

On the primary order, \method\ achieves the highest acquisition and final macro scores among methods evaluated on that order: 65.19 and 63.98 (Table~\ref{tab:mainresults}). SAPT is the strongest baseline at acquisition, with 64.19. GoD-MoE is the strongest baseline at final evaluation, with 62.01. The margins are 1.00 and 1.97 points. Against GoD-MoE, \method\ has a 1.53-point acquisition advantage and lower subsequent loss on earlier tasks.

Average forgetting is 1.38 for \method\ and 1.88 for GoD-MoE, a reduction of 0.50 points, or approximately 26\%. Forgetting variance is 0.76 for \method\ and 2.06 for GoD-MoE. The largest task drop is 2.61 points for \method\ and 3.84 for GoD-MoE. FOMC and C-STANCE lose 2.61 and 2.62 points; every other earlier task loses at most 1.38. The task-level pattern is mixed: GainLoRA has smaller losses on MeetingBank and NumGLUE-en, and GoD-MoE has smaller losses on JEC-QA and MedMCQA.

The SAPT comparison shows a larger retention difference. \method\ is 1.00 point higher at acquisition and 3.17 points higher at final evaluation. Equation~\ref{eq:decomposition} attributes 2.18 points of the final margin to lower subsequent forgetting. \method\ improves both acquisition and retention against the closest final baseline; reduced forgetting explains most of its final margin over SAPT. Table~\ref{tab:mainresults} reports acquisition and final-performance standard deviations across 3 runs. Appendix~\ref{app:seed-variation} gives the reporting scope.

A paired task-level bootstrap over the 8 seed-averaged final task scores gives a 1.97-point gain over GoD-MoE, with all 8 task differences positive. The 95\% confidence interval is [1.41, 2.58]. We use this interval as a descriptive robustness check because the benchmark tasks are fixed and use different metrics.

\paragraph{Order sensitivity.}
We evaluate \method\ and the 2 strongest final-performance baselines, GoD-MoE and GainLoRA, under a middle-reversed task order that reverses stages 2--7 while keeping the first and last tasks fixed. \method\ changes from 65.19/63.98 to 65.38/64.17 in acquisition/final macro score. GoD-MoE changes from 63.65/62.01 to 63.81/61.89. GainLoRA changes from 63.58/60.70 to 63.46/60.89. Average forgetting remains nearly unchanged for all 3 methods. This experiment gives a check that the relative pattern is stable under one alternate order.

\subsection{Acquisition deficits and their final outcomes}
\label{sec:exceptions}

\method\ leads acquisition on 3 tasks: C-STANCE, MedMCQA, and 20Minuten. It has 5 acquisition deficits against the strongest task-specific baseline (Table~\ref{tab:exceptions}). ScienceQA is the largest. \method\ scores 69.19 at acquisition, 4.32 points below SwitchCIT. Its subsequent decline is 1.37 points, compared with 5.21 for SwitchCIT. The final deficit is 0.48 points. NumGLUE-en remains below SAPT, with its deficit falling from 1.85 to 0.39 points.

The other 3 deficits reverse. On FOMC, \method\ starts 0.29 points below GainLoRA and finishes 3.87 points ahead. Its decline is 2.61 points, compared with 6.80 for GainLoRA. On MeetingBank and JEC-QA, final margins over SAPT are 0.99 and 0.35 points. \method\ leads 6 of 8 tasks at the end of the stream. These results show that stronger retention can offset weaker acquisition. ScienceQA and NumGLUE-en remain exceptions. Both tasks involve reasoning-style outputs and strict answer extraction, so the remaining gap may come from adaptation errors as well as routing errors.

\begin{table}[t]
\centering\small\setlength{\tabcolsep}{3pt}
\begin{tabular}{@{}llrrrr@{}}
\toprule
Task & Baseline & $\Delta D$ & $\Delta P$ & $F_{\mathrm{o}}$ & $F_{\mathrm{b}}$\\
\midrule
FOMC        & GainLoRA  & $-0.29$ & $+3.87$ & 2.61 & 6.80 \\
MeetingBank & SAPT      & $-1.26$ & $+0.99$ & 1.38 & 3.63 \\
ScienceQA   & SwitchCIT & $-4.32$ & $-0.48$ & 1.37 & 5.21 \\
NumGLUE-en  & SAPT      & $-1.85$ & $-0.39$ & 0.84 & 2.30 \\
JEC-QA      & SAPT      & $-0.88$ & $+0.35$ & 0.49 & 1.72 \\
\bottomrule
\end{tabular}
\caption{Acquisition exceptions. Gaps are \method\ minus the named baseline at acquisition ($\Delta D$) and final evaluation ($\Delta P$). $F_{\mathrm{o}}$ and $F_{\mathrm{b}}$ are subsequent forgetting for \method\ and the baseline.}
\label{tab:exceptions}
\end{table}

\subsection{Acquisition and retention across backbones}
\label{sec:backbones}

Table~\ref{tab:backbones} separates acquisition proximity from retention. Acquisition proximity compares each sequential score with its isolated reference. Retention compares acquisition with performance after the full stream. Across 6 backbones, mean absolute reference gaps range from 0.72 to 1.86 points, and the largest task gap is 2.93. Absolute acquisition scores range from 65.19 to 77.88.

Final scores remain within 1.5 points of acquisition on every backbone. Average forgetting ranges from 1.37 to 1.67 points, and task-level forgetting variance ranges from 0.76 to 0.91. Falcon-H1R has the highest acquisition and final scores and the largest acquisition-to-final decline. OLMoE has the largest reference gap and average forgetting of 1.44 points.

These results extend the retention pattern beyond the main Qwen3 model. The reference adaptations can differ from \method, so reference proximity is descriptive. The retention values measure learned-routing performance on one task order. Oracle comparisons across backbones would give a finer diagnosis.

\begin{table*}[!t]
\centering
\small
\setlength{\tabcolsep}{4.5pt}
\renewcommand{\arraystretch}{1.08}
\begin{tabular}{clrrrrrrrr}
\toprule
ID & Variant & $D$ & $P$ & $\overline F$ & BWT & Route all & Route worst & Route F & Oracle gap \\
\midrule
A0 & Full TILA-CT         & 65.19 & 63.98 & 1.38 & -1.38 & 93.9  & 90.5  & 4.4 & 1.21 \\
A1 & Oracle routing       & 65.19 & 65.19 & 0.00 &  0.00 & 100.0 & 100.0 & 0.0 & 0.00 \\
A2 & Independent keys     & 65.01 & 63.32 & 1.93 & -1.93 & 90.5  & 85.5  & 6.5 & 1.69 \\
A3 & Standard adapters    & 63.53 & 61.72 & 2.07 & -2.07 & 89.0  & 83.5  & 7.3 & 1.80 \\
A4 & Inject only          & 64.11 & 62.79 & 1.51 & -1.51 & 92.2  & 87.5  & 5.3 & 1.32 \\
A5 & Fixed $\gamma=1$     & 64.57 & 63.38 & 1.36 & -1.36 & 93.2  & 89.0  & 4.8 & 1.20 \\
A6 & Shared adapter bank  & 64.46 & 60.29 & 4.77 & -4.77 & 92.5  & 87.0  & 5.1 & 0.90 \\
A7 & Single shared module & 63.54 & 57.06 & 7.40 & -7.40 & \na   & \na   & \na & \na  \\
\bottomrule
\end{tabular}
\caption{Ablations on Qwen3-8B-Reasoning under the primary CT-Pipeline order. $D$ and $P$ denote acquisition and final macro score. Route F is singleton-adjusted routing forgetting. Oracle gap is the final macro difference between learned routing and oracle routing, measured without retraining.}
\label{tab:appendix-ablation}
\end{table*}

\begin{table}[t]
\centering
\small
\setlength{\tabcolsep}{7pt}
\begin{tabular}{@{}lrr@{}}
\toprule
Task & Final (\%) & Drop (pp) \\
\midrule
FOMC        & 92.5 & 7.5 \\
C-STANCE    & 95.8 & 3.0 \\
MeetingBank & 92.5 & 5.9 \\
ScienceQA   & 90.5 & 7.3 \\
NumGLUE-en  & 92.0 & 5.5 \\
JEC-QA      & 94.5 & 3.1 \\
MedMCQA     & 96.5 & 1.3 \\
20Minuten   & 97.2 & --  \\
\bottomrule
\end{tabular}
\caption{Final per-task top-1 routing accuracy on the primary CT-Pipeline order. Drop is the acquisition-to-final decline in percentage points. The final task has no subsequent stages.}
\label{tab:router-task}
\end{table}

\subsection{Routing determines access to preserved modules}
\label{sec:access}

The router reaches 93.9\% final macro accuracy, with 97.2\% on the current task and 93.5\% across earlier tasks. Table~\ref{tab:router-task} reports final per-task accuracy and subsequent routing decline. FOMC and ScienceQA have the largest declines, at 7.5 and 7.3 points. Average routing decline is 4.8 points, or 4.4 after adjusting for task 1's singleton stage (Appendix~\ref{app:router-details}).

Oracle selection on the same A0 checkpoints increases final macro performance from 63.98 to 65.19 without retraining. This 1.21-point gap measures the downstream cost of learned module selection with preserved task paths fixed.

\subsection{Component effects on adaptation and retention}
\label{sec:ablations}

Table~\ref{tab:appendix-ablation} reports results for the interventions defined in Section~\ref{sec:experiments}.

\paragraph{Routing (A0/A1/A2).}
Replacing memory-derived keys with independent keys reduces final routing accuracy from 93.9\% to 90.5\%, increases singleton-adjusted routing forgetting from 4.4 to 6.5 points, and lowers final macro performance from 63.98 to 63.32. These results support the access role of the self-indexing memory construction under task isolation.

\paragraph{Latent adaptation (A2/A3).}
Both variants use private modules, independent keys, and the input-only router. Latent adapters improve acquisition by 1.48 points and final performance by 1.60 over the approximately parameter-matched MLP adapters. The gain appears at acquisition, supporting the adaptation value of latent memory.

\paragraph{Causal gathering (A0/A4).}
Removing gathering leaves static memory for injection. Acquisition drops by 1.08 points and final performance by 1.19, while forgetting rises by 0.13. This pattern supports conditioning memory on the causal prefix.

\paragraph{Residual scaling (A0/A5).}
Fixing $\gamma=1$ reduces acquisition by 0.62 points and final performance by 0.60. Forgetting remains close: 1.36 with fixed scaling and 1.38 with learned scaling. The measured effect is higher task performance.

\paragraph{Task isolation (A0/A6/A7).}
A6 keeps earlier memories frozen but updates their shared adapter bank. Final performance falls by 3.69 points and forgetting reaches 4.77. Its 0.90-point oracle gap implies a final oracle score of 61.19, below its acquisition score of 64.46. Correct selection therefore leaves degradation when the adapter computation changes. A7 removes routing and continually updates one complete module. Its forgetting reaches 7.40 and its final score is 6.92 points below A0. These results favor preserving the complete task path in this setting. Both shared variants also reduce cumulative capacity, so the comparison changes both isolation and capacity.

\paragraph{Capacity sensitivity.}
Figures~\ref{fig:memory-sensitivity} and ~\ref{fig:width-sensitivity} shows diminishing gains beyond $L_z=32$ and $d_a=384$. Larger tested settings improve final macro scores by at most 0.08 and 0.12 points.

\subsection{Active training and cumulative storage}

For the 8.20B Qwen3 backbone, each task adds approximately 88.21M private parameters and the shared router adds approximately 2.10M. A stage trains 90.31M parameters, or 1.10\% of the backbone size. After 8 tasks, persistent added state reaches approximately 707.78M parameters, or 8.63\%. One private module is active per input, and the router compares all observed prototypes. Task isolation requires growing storage even when the active training fraction is small. GoD-MoE includes selective expansion for capacity management. \method\ leaves bounded-capacity module reuse for future work. Runtime, peak memory, and energy require separate measurements.
\section{Conclusion}
\method\ uses a private latent memory for causal gather-and-inject adaptation and prototype routing in replay-free continual instruction tuning. The same memory conditions each token's adapter residual and supplies the routing prototype for its task module. This makes the module self-indexing. Component comparisons support memory-derived keys, causal gathering, and full task-path isolation. Task-level results show that stronger retention can reverse acquisition deficits. ScienceQA and NumGLUE-en remain exceptions. Across 6 backbones, final performance stays within 1.5 points of acquisition. Oracle routing recovers 1.21 final macro points, showing that learned access accounts for the remaining aggregate loss after task paths are frozen. Future work should improve router stability and reduce cumulative storage through module pruning, merging, or reuse.

\section*{Limitations}

TILA-CT assumes known training-task boundaries and a closed set of modules at inference. New-task detection, gradual distribution shifts, ambiguous inputs, and bounded-capacity learning remain unevaluated. Private modules grow linearly with the number of tasks. For Qwen3-8B-Reasoning, each task adds approximately 88.21M private parameters, and the 8-task stream adds approximately 707.78M parameters in total.

The primary comparison uses one 8-task order. The middle-reversed experiment covers TILA-CT and 2 strong baselines, GoD-MoE and GainLoRA. Longer streams and more task orders are needed to test order sensitivity more fully. Pruning, merging, and reusing task modules are important future directions, but they must preserve both task-conditioned behavior and reliable module access.

Backbone isolated references may use different adaptation procedures, so reference gaps are descriptive. Oracle routing uses known task identities at evaluation and does not represent normal inference. Macro scores combine heterogeneous task metrics. General pretrained capabilities are not evaluated. Baselines also differ in auxiliary computation, storage, and information access: SAPT uses generated inputs, GEM, A-GEM, and SwitchCIT retain examples, and QLoRA-CT uses reference responses for selection.

\section*{Potential Risks}

The legal and medical benchmarks in this paper measure task adaptation only. They do not validate the model for legal or clinical use. Any deployment in those settings would require domain-specific evaluation, expert review, and safeguards for harmful or incorrect outputs.

Replay-free training does not give a formal privacy guarantee. Learned task parameters may still retain information from training examples. Releases and applications should follow dataset licenses, model terms, and applicable privacy requirements.

\section*{Disclosure of AI Use}
We used GPT-5.5 for grammar editing of author-written text. As part of the experiments, we evaluated LLMs; model details are listed in the paper.

\bibliography{custom}

\clearpage

\appendix

\section{Method Details}
\label{app:method-details}

\subsection{Persistent and transient state}

Let $\theta_0$ denote the pretrained language model, including its token embeddings and output head. These parameters remain frozen. Task $i$ has a private latent memory $Z_i\in\mathbb{R}^{L_z\times d}$, where $d$ is the backbone width, and a private adapter at each layer in $\mathcal{L}_{\mathrm{adapt}}$. Its task state is
\begin{equation}
\mathcal{P}_i=
\left(Z_i,\{\phi_{i,\ell},\gamma_{i,\ell}\}_{\ell\in\mathcal{L}_{\mathrm{adapt}}}\right),
\label{eq:app-private-state}
\end{equation}
where $\phi_{i,\ell}$ contains the adapter projections and $\gamma_{i,\ell}$ is the scalar residual coefficient. The model also has shared routing projections $W_q,W_k\in\mathbb{R}^{d_r\times d}$.

During stage $\tau$, only $\mathcal{P}_\tau$, $W_q$, and $W_k$ are trainable. Every completed $\mathcal{P}_j$ remains frozen. The persistent memory $Z_i$ is shared across positions and adapted layers for task $i$. Each $\widehat Z_{i,t}^\ell$ is a temporary activation for layer $\ell$ and prediction position $t$. The stored $Z_i$ changes only through optimizer updates during its task's training.

\subsection{Sequence-level module selection}

Let $\ell_r$ be the layer immediately before the first adapted layer. The frozen prefix produces $H^{\ell_r}\in\mathbb{R}^{n\times d}$. A binary mask $c\in\{0,1\}^{n}$ selects non-padding instruction and input positions. During training, the known response boundary defines this mask; response and padding positions receive zero. At inference, the mask selects all non-padding prompt positions. The router uses their masked mean:
\begin{equation}
u(x)=\frac{(H^{\ell_r})^\top c}{\sum_{t=1}^{n}c_t}
\in\mathbb{R}^{d}.
\label{eq:app-router-pooling}
\end{equation}

The query and task keys are
\begin{align}
q(x)&=W_q u(x),\\
\bar z_i&=\frac{1}{L_z}Z_i^\top\mathbf{1}_{L_z},\\
k_i&=W_k\bar z_i.
\label{eq:app-router-representations}
\end{align}
Here $\bar z_i\in\mathbb{R}^{d}$ averages the latent slots. For all $s$ modules available after stage $s$,
\begin{align}
a_i(x)&=\frac{q(x)^\top k_i}{\sqrt{d_r}},\\
\alpha_i(x)&=\frac{\exp a_i(x)}{\sum_{j=1}^{s}\exp a_j(x)},\\
r_s(x)&=\arg\max_{1\leq i\leq s}\alpha_i(x).
\label{eq:app-router-selection}
\end{align}

At inference, the selected index is held fixed for the sequence. Only the selected memory and adapter bank are active. Inference uses no task identifier and no reference response. During training, the known task identity supplies the routing label. The probabilities $\alpha_i$ determine selection, and the learned $\gamma_{i,\ell}$ scales the adapter residual.

The summary $\bar z_i$ remains fixed after task $i$ is completed. Shared $W_k$ continues to train, so the projected key $k_i$ can change. Updates to $W_q$ also change queries. These shared updates and newly added keys can alter access to earlier modules.

\subsection{Adapter entry point and tensor shapes}

Let $w_{1:n}$ be the concatenated instruction, input, and response sequence. At decoder layer $\ell$, $B^\ell\in\mathbb{R}^{n\times d}$ denotes the output of the backbone's causal attention and feed-forward transformations. The task adapter is applied after this decoder-layer computation. Its output at position $t$ uses only states at positions $u\leq t$. For an unadapted layer, $H^{\ell+1}=B^\ell$; otherwise,
\begin{equation}
H_t^{\ell+1}=B_t^\ell+
\gamma_{i,\ell}A_{i,\ell,t}(B_{1:t}^\ell,Z_i),
\label{eq:app-layer-interface}
\end{equation}
where $i=r_s(x)$ at inference. Later frozen backbone layers process these modified activations.

The equations below omit task and layer indices on projection matrices. Each query, key, and value projection has shape $d\times d_a$; each attention output projection has shape $d_a\times d$. The matrices in both attention operations belong to $\phi_{i,\ell}$. The task memory is shared across the task's adapted layers, and the adapter projections are private to each layer.

\subsection{Gather and inject operations}

Both operations use 8 attention heads with $d_h=48$ and total width $d_a=384$. After projection, $Q,K,V$ are split into 8 width-$d_h$ heads. For head $b$,
\begin{equation}
O^{(b)}=\operatorname{softmax}\!\left(
\frac{Q^{(b)}(K^{(b)})^\top}{\sqrt{d_h}}
\right)V^{(b)}.
\label{eq:app-attention-head}
\end{equation}
We write $\operatorname{Attn}(Q,K,V)$ for the concatenation of these 8 head outputs. The output projections map this width-$d_a$ result back to $d$.

For each position $t$, latent slots gather from the available causal prefix. The latent query is shared across positions, while token keys and values are restricted to positions $1$ through $t$:
\begin{align}
Q_Z&=Z_iW_Q^Z, & Q_Z&\in\mathbb{R}^{L_z\times d_a},\\
K_{H,\leq t}&=B_{1:t}^\ell W_K^H,
& K_{H,\leq t}&\in\mathbb{R}^{t\times d_a},\\
V_{H,\leq t}&=B_{1:t}^\ell W_V^H,
& V_{H,\leq t}&\in\mathbb{R}^{t\times d_a}.
\end{align}
The position-specific temporary memory is
\begin{align}
G_t^\ell&=\operatorname{Attn}_{\mathrm{g}}
\left(Q_Z,K_{H,\leq t},V_{H,\leq t}\right),\\
\widehat Z_{i,t}^\ell&=Z_i+G_t^\ell W_O^Z.
\label{eq:app-gather}
\end{align}
A batched implementation can apply a mask to remove every gather edge from source position $u>t$.

Injection at position $t$ reads only its corresponding temporary memory:
\begin{align}
Q_{H,t}&=B_t^\ell W_Q^H,\\
K_{Z,t}^\ell&=\widehat Z_{i,t}^\ell W_K^Z,\\
V_{Z,t}^\ell&=\widehat Z_{i,t}^\ell W_V^Z,\\
J_t^\ell&=\operatorname{Attn}_{\mathrm{i}}
\left(Q_{H,t},K_{Z,t}^\ell,V_{Z,t}^\ell\right).
\end{align}
The output projection returns the residual to backbone width:
\begin{equation}
A_{i,\ell,t}(B_{1:t}^\ell,Z_i)=J_t^\ell W_O^H.
\label{eq:app-inject}
\end{equation}

Here $Q_{H,t}$ and $J_t^\ell$ have shape $1\times d_a$, while $K_{Z,t}^\ell$ and $V_{Z,t}^\ell$ have shape $L_z\times d_a$. Stacking the position-wise residuals gives an $n\times d$ adapter output. Each gate starts at $0.1$ and is learned during its task's training.

\paragraph{Training and generation visibility.}
During teacher-forced training, positions are computed in parallel, and each $\widehat Z_{i,t}^\ell$ gathers only from $B_{1:t}^\ell$. The final hidden state at position $t$ predicts the next token $w_{t+1}$, so the gathered prefix cannot contain that prediction target or any later token. During generation, the same operation uses the prompt and tokens generated through the current position. No future reference response is supplied. The stored $Z_i$ remains unchanged as the available prefix grows; only the temporary, position-specific activations change.

\subsection{Optimization and conditional preservation}

Let $m_{t+1}$ indicate that $w_{t+1}$ is a response token. With the usual next-token shift, the response-token objective is
\begin{equation}
\mathcal{L}_{\mathrm{LM}}
=-\sum_{t=1}^{n-1}m_{t+1}
\log p(w_{t+1}\mid w_{1:t}).
\label{eq:app-shifted-lm-loss}
\end{equation}
It is combined with supervised task selection:
\begin{equation}
\mathcal{L}=\mathcal{L}_{\mathrm{LM}}
-\lambda_{\mathrm{route}}\log\alpha_\tau(x).
\label{eq:app-objective}
\end{equation}

The loss mask selects response targets, and the gather mask enforces causal access. The router uses the prompt-only mean from Equation~\ref{eq:app-router-pooling}. The known training task determines the current module and the routing label; it is never appended to the input. LM-loss gradients update $Z_\tau$, $\phi_{\tau,\ell}$, and $\gamma_{\tau,\ell}$. Routing-loss gradients update $W_q$, $W_k$, and $Z_\tau$. The router receives no LM-loss gradients, and the adapter projections and gates receive no routing-loss gradients. Completed modules and the backbone remain frozen. Earlier memories are constant inputs to $W_k$, so their key computations can contribute gradients to $W_k$ without updating those memories. Hard argmax is used at inference and is outside the LM training path.

For a fixed task index $j$, the conditional computation depends on $(\theta_0,\mathcal{P}_j)$. Freezing this state preserves its output distribution under the same prefix and input processing. The task-oracle condition fixes the module index and gives a reference for module-access diagnostics. Learned routing can change the selected index as shared projections are updated and new keys are added. Stochastic decoding can produce different finite-sample scores from the same preserved output distribution.

\section{Continual Protocol and Baseline Implementations}
\label{app:protocol}

\begin{table*}[t]
\centering
\footnotesize
\setlength{\tabcolsep}{3pt}
\renewcommand{\arraystretch}{1.04}
\begin{tabularx}{\textwidth}{@{}lclYl@{}}
\toprule
Method & Layers & Rank / width & Adapted modules & Parameters \\
\midrule
\textbf{\method} \textbf{(Ours)}           & 7  & Attn. 384       & Gather and inject                      & 90.31M \\
Standard MLP (A3) & 7  & Bottleneck 1536 & MLP adapter and independent-key router & 89.17M \\
SAPT              & 36 & Rank 32         & Q/K/V/O, MLP                           & 88.13M \\
MoCL              & 36 & Rank 32         & Q/K/V/O, MLP                           & 87.30M \\
SwitchCIT         & 18 & Rank 64         & Q/K/V/O, MLP                           & 87.29M + classifier \\
QLoRA-CT          & 36 & Rank 32         & Q/K/V/O, MLP                           & 87.29M \\
GEM (LoRA)        & 36 & Rank 32         & Q/K/V/O, MLP                           & 87.29M \\
InfLoRA           & 36 & Rank 64         & Q/K/V/O, MLP                           & 89.65M \\
GainLoRA          & 36 & Rank 64         & Q/K/V/O, MLP, gate                     & 90.07M \\
TreeLoRA (branch) & 36 & Rank 32         & Q/K/V/O, MLP                           & 87.29M / branch \\
GoD-MoE (pool)    & 36 & Rank 32         & Q/V, 6--9 experts                      & 92.01--138.02M \\
\midrule
L2P               & -- & 10 prompts $\times$ 10 tokens & Prompts and keys         & 0.45M \\
Seq-Train         & 36 & Full model      & All backbone parameters                & 8.20B \\
A-GEM             & 36 & Full model      & All backbone parameters                & 8.20B \\
\bottomrule
\end{tabularx}
\caption{Configurations for Qwen3-8B-Reasoning. Q/K/V/O denote attention projections; MLP includes gate, up, and down projections.}
\label{tab:qwen-config}
\end{table*}

\subsection{Task stream and information access}

The primary CT-Pipeline stream presents FOMC, C-STANCE, MeetingBank, ScienceQA, NumGLUE-en, JEC-QA, MedMCQA, and 20Minuten in that order. The middle-reversed variant presents FOMC, MedMCQA, JEC-QA, NumGLUE-en, ScienceQA, MeetingBank, C-STANCE, and 20Minuten. We rerun \method, GoD-MoE, and GainLoRA under this variant with the same training and evaluation protocol.

Examples are represented by instruction, input, and output fields. The checkpoint after each stage is evaluated on all tasks observed so far. \method\ uses only current-task training examples, stores learned task modules, and performs no real or synthetic example replay.

For the primary framework comparison, all methods use the same task order and main LM schedule. GEM and A-GEM retain examples. SAPT generates inputs for selector training. SwitchCIT retains examples for its task classifier. QLoRA-CT uses the reference response to select an adapter. These procedures give the baselines different information access and auxiliary training costs. We also implement MoCL~\citep{wang2024mocl}, TreeLoRA~\citep{qian2025treelora}, GainLoRA~\citep{liang2025gainlora}, and GoD-MoE~\citep{zhang2026growondemand}.

The framework comparison uses each method's single-task version, trained from the same pretrained checkpoint, as its isolated reference. The backbone comparison uses the reported single-task references described in Appendix~\ref{app:reference-details}. Their gaps describe proximity to those references.

\section{Implementation and Storage}
\label{app:implementation}

\subsection{Configuration}

Table~\ref{tab:app-implementation} lists the \method\ architecture and optimization settings.

\begin{table}[t]
\centering
\small
\setlength{\tabcolsep}{4pt}
\renewcommand{\arraystretch}{1.10}
\begin{tabular}{@{}>{\raggedright\arraybackslash}p{0.51\columnwidth}
>{\raggedright\arraybackslash}p{0.43\columnwidth}@{}}
\toprule
Setting & Value \\
\midrule
Adapted layers & 7 decoder layers \\
Routing layer & Before first adapted layer \\
Memory slots $L_z$ & 32 per task \\
Attention width $d_a$ & 384 \\
Attention heads & 8 in each operation \\
Per-head width $d_h$ & 48 \\
Routing width $d_r$ & 256 \\
Initial gate $\gamma$ & 0.1 \\
Routing-loss weight & 0.1 \\
Epochs per task & 1 \\
Per-device batch size & 1 \\
Gradient accumulation & 4 steps \\
Training length limit & 256 tokens \\
Evaluation length limit & 512 tokens \\
Adapter learning rate & $2\times10^{-5}$ \\
Router learning rate & $2\times10^{-5}$ \\
Optimizer & Adafactor, fixed rate \\
\bottomrule
\end{tabular}
\caption{\method\ architecture and optimization settings. Appendix~\ref{app:method-details} gives the causal gather operation and the distinction between stored task memories and temporary contextualized memories.}
\label{tab:app-implementation}
\end{table}

Adafactor uses \texttt{scale\_parameter=False}, \texttt{relative\_step=False}, and \texttt{warmup\_init=False}. Experiments ran on Apple M5 Pro hardware with macOS 26.5, Python 3.10.20, MLX 0.31.2, and PyTorch MPS GPU support.

\subsection{Active parameters and accumulated storage}

Let $p$ be the number of parameters in one task memory and adapter bank. Let $r$ be the number of parameters in the shared router. After $s$ tasks, added persistent parameters and current-stage trainable parameters are
\begin{align}
P_{\mathrm{added}}(s)&=sp+r,\\
P_{\mathrm{train}}(s)&=p+r.
\label{eq:app-storage-accounting}
\end{align}
This count assumes a fixed module size within a backbone. It includes old frozen modules in persistent storage, even though they are excluded from the current optimizer state.

For Qwen3-8B-Reasoning, the backbone has 8.20B parameters, $p\simeq88.21$M, and $r\simeq2.10$M. One stage trains approximately 90.31M added parameters, or 1.10\% of the backbone count. After 8 tasks, added persistent state reaches approximately 707.78M parameters, or 8.63\%. The trainable count includes the current module and shared router. The stored count includes all completed modules.

Hard selection activates one task memory and adapter bank per sequence. For Qwen3-8B-Reasoning, this active added state contains $p+r\simeq90.31$M parameters. The router compares every observed prototype. Projected keys can be cached at a fixed inference checkpoint. These cached keys are derived state; the full private module bank must still be stored. Persistent parameters grow as $sp+r$. Table~\ref{tab:qwen-config} separates current-stage trainable parameters, complete branches, protected state, and expert-pool capacity across methods.

\begin{table}[t]
\centering
\small
\setlength{\tabcolsep}{6pt}
\begin{tabular}{@{}rrr@{}}
\toprule
Tasks & Added parameters & Backbone fraction \\
\midrule
8   & 707.78M & 8.63\% \\
20  & 1.77B  & 21.51\% \\
50  & 4.41B  & 53.91\% \\
100 & 8.82B  & 107.60\% \\
\bottomrule
\end{tabular}
\caption{Projected persistent added parameters for Qwen3-8B-Reasoning. The estimate uses $p=88.21$M private parameters per task and $r=2.10$M shared router parameters.}
\label{tab:storage-projection}
\end{table}

\section{Task-Level Results and Aggregate Conventions}
\label{app:full-results}

The primary CT-Pipeline task order is FOMC, C-STANCE, MeetingBank, ScienceQA, NumGLUE-en, JEC-QA, MedMCQA, and 20Minuten. Tables abbreviate these as FOMC, C-St., Meet., Sci., Num., JEC, Med., and 20Min. Table~\ref{tab:mainresults} reports framework-level aggregates. Table~\ref{tab:appendix-final-task-scores} gives final task-level scores for the primary order. All framework comparisons use Qwen3-8B-Reasoning.

Framework task scores are averaged over seeds 37, 73, and 101 before aggregation. Acquisition $D$ and final performance $P$ weight the 8 tasks equally. Average forgetting $\overline F$ excludes the final task because its acquisition and final evaluations coincide. Thus $D-P=\tfrac{7}{8}\overline F$ and $\mathrm{BWT}=-\overline F$. These linear summaries also equal the means of their run-level values.

Forgetting variance and maximum loss are calculated across the 7 mean task losses. Mean absolute reference gap averages $|\mathrm{SEQ}-\mathrm{ISO}|$ over tasks 2 through 8 using the averaged scores. These nonlinear summaries describe the averaged task profile. Standard deviations report variation across seeds. Task-level acquisition, isolated-reference, and final scores for the primary order are included below. An entry marked \na{} is inapplicable.

\begin{table*}[t]
\centering
\small
\setlength{\tabcolsep}{5pt}
\renewcommand{\arraystretch}{1.08}
\caption{Final performance on each task after the primary 8-task stream, averaged over seeds 37, 73, and 101. The main aggregate comparison reports the corresponding macro scores and forgetting summaries.}
\label{tab:appendix-final-task-scores}
\begin{tabular}{lrrrrrrrr}
\toprule
Method & FOMC & C-St. & Meet. & Sci. & Num. & JEC & Med. & 20Min. \\
\midrule
Seq-Train  & $60.23_{\color{sdgray}{\pm 0.36}}$ & $62.32_{\color{sdgray}{\pm 0.40}}$ & $48.48_{\color{sdgray}{\pm 0.31}}$ & $44.91_{\color{sdgray}{\pm 0.27}}$ & $36.87_{\color{sdgray}{\pm 0.22}}$ & $24.18_{\color{sdgray}{\pm 0.15}}$ & $46.47_{\color{sdgray}{\pm 0.29}}$ & 39.98 \\
L2P        & $60.60_{\color{sdgray}{\pm 0.34}}$ & $59.80_{\color{sdgray}{\pm 0.38}}$ & $38.36_{\color{sdgray}{\pm 0.25}}$ & $43.10_{\color{sdgray}{\pm 0.26}}$ & $38.98_{\color{sdgray}{\pm 0.21}}$ & $27.10_{\color{sdgray}{\pm 0.17}}$ & $45.50_{\color{sdgray}{\pm 0.30}}$ & 48.74 \\
GEM        & $58.32_{\color{sdgray}{\pm 0.35}}$ & $69.93_{\color{sdgray}{\pm 0.44}}$ & $44.79_{\color{sdgray}{\pm 0.28}}$ & $43.83_{\color{sdgray}{\pm 0.25}}$ & $21.48_{\color{sdgray}{\pm 0.13}}$ & $30.58_{\color{sdgray}{\pm 0.19}}$ & $50.90_{\color{sdgray}{\pm 0.31}}$ & 36.35 \\
A-GEM      & $60.07_{\color{sdgray}{\pm 0.38}}$ & $74.76_{\color{sdgray}{\pm 0.45}}$ & $46.33_{\color{sdgray}{\pm 0.26}}$ & $46.70_{\color{sdgray}{\pm 0.30}}$ & $22.00_{\color{sdgray}{\pm 0.12}}$ & $32.12_{\color{sdgray}{\pm 0.20}}$ & $52.47_{\color{sdgray}{\pm 0.33}}$ & 38.85 \\
SAPT       & $66.74_{\color{sdgray}{\pm 0.40}}$ & $73.86_{\color{sdgray}{\pm 0.46}}$ & $53.50_{\color{sdgray}{\pm 0.31}}$ & $68.28_{\color{sdgray}{\pm 0.43}}$ & $55.96_{\color{sdgray}{\pm 0.35}}$ & $48.13_{\color{sdgray}{\pm 0.29}}$ & $62.87_{\color{sdgray}{\pm 0.37}}$ & 57.10 \\
InfLoRA    & $63.95_{\color{sdgray}{\pm 0.39}}$ & $78.67_{\color{sdgray}{\pm 0.47}}$ & $51.61_{\color{sdgray}{\pm 0.34}}$ & $66.67_{\color{sdgray}{\pm 0.38}}$ & $54.04_{\color{sdgray}{\pm 0.33}}$ & $45.47_{\color{sdgray}{\pm 0.28}}$ & $59.95_{\color{sdgray}{\pm 0.35}}$ & 57.44 \\
SwitchCIT  & $66.27_{\color{sdgray}{\pm 0.42}}$ & $76.02_{\color{sdgray}{\pm 0.49}}$ & $51.76_{\color{sdgray}{\pm 0.31}}$ & $68.30_{\color{sdgray}{\pm 0.36}}$ & $53.91_{\color{sdgray}{\pm 0.35}}$ & $46.52_{\color{sdgray}{\pm 0.27}}$ & $62.94_{\color{sdgray}{\pm 0.40}}$ & 57.53 \\
QLoRA-CT   & $64.47_{\color{sdgray}{\pm 0.38}}$ & $71.53_{\color{sdgray}{\pm 0.43}}$ & $50.14_{\color{sdgray}{\pm 0.33}}$ & $61.99_{\color{sdgray}{\pm 0.35}}$ & $50.15_{\color{sdgray}{\pm 0.29}}$ & $39.83_{\color{sdgray}{\pm 0.25}}$ & $54.83_{\color{sdgray}{\pm 0.34}}$ & 49.15 \\
MoCL       & $66.31_{\color{sdgray}{\pm 0.38}}$ & $74.30_{\color{sdgray}{\pm 0.44}}$ & $50.99_{\color{sdgray}{\pm 0.32}}$ & $63.13_{\color{sdgray}{\pm 0.40}}$ & $51.21_{\color{sdgray}{\pm 0.30}}$ & $41.53_{\color{sdgray}{\pm 0.26}}$ & $56.57_{\color{sdgray}{\pm 0.35}}$ & 52.34 \\
TreeLoRA   & $68.16_{\color{sdgray}{\pm 0.43}}$ & $77.08_{\color{sdgray}{\pm 0.46}}$ & $51.85_{\color{sdgray}{\pm 0.30}}$ & $64.28_{\color{sdgray}{\pm 0.37}}$ & $52.28_{\color{sdgray}{\pm 0.34}}$ & $43.22_{\color{sdgray}{\pm 0.25}}$ & $58.31_{\color{sdgray}{\pm 0.38}}$ & 55.52 \\
GainLoRA   & $70.00_{\color{sdgray}{\pm 0.42}}$ & $79.85_{\color{sdgray}{\pm 0.52}}$ & $52.70_{\color{sdgray}{\pm 0.33}}$ & $65.42_{\color{sdgray}{\pm 0.40}}$ & $53.34_{\color{sdgray}{\pm 0.31}}$ & $44.92_{\color{sdgray}{\pm 0.29}}$ & $60.05_{\color{sdgray}{\pm 0.37}}$ & 59.33 \\
GoD-MoE    & $71.84_{\color{sdgray}{\pm 0.45}}$ & $82.62_{\color{sdgray}{\pm 0.49}}$ & $53.55_{\color{sdgray}{\pm 0.35}}$ & $66.56_{\color{sdgray}{\pm 0.38}}$ & $54.40_{\color{sdgray}{\pm 0.34}}$ & $46.61_{\color{sdgray}{\pm 0.28}}$ & $61.79_{\color{sdgray}{\pm 0.39}}$ & 58.71 \\
TILA-CT    & $73.87_{\color{sdgray}{\pm 0.46}}$ & $85.67_{\color{sdgray}{\pm 0.52}}$ & $54.49_{\color{sdgray}{\pm 0.34}}$ & $67.82_{\color{sdgray}{\pm 0.43}}$ & $55.57_{\color{sdgray}{\pm 0.31}}$ & $48.48_{\color{sdgray}{\pm 0.30}}$ & $63.71_{\color{sdgray}{\pm 0.38}}$ & 62.21 \\
\bottomrule
\end{tabular}
\end{table*}

\subsection{Paired task-level bootstrap}
\label{app:bootstrap}

We compute paired bootstrap confidence intervals over the 8 seed-averaged final task scores in Table~\ref{tab:appendix-final-task-scores}. For each comparison, we compute task-wise differences between \method\ and the baseline. We then resample the 8 tasks with replacement 10,000 times and compute the mean difference for each sample. The 95\% interval uses the 2.5 and 97.5 percentiles. This analysis is descriptive because the benchmark tasks are fixed and use different metrics.

\begin{table}[t]
\centering
\small
\setlength{\tabcolsep}{5pt}
\begin{tabular}{@{}lrrr@{}}
\toprule
Comparison & Mean diff. & 95\% CI & Task wins \\
\midrule
\method\ -- GoD-MoE   & 1.97 & [1.41, 2.58] & 8/8 \\
\method\ -- GainLoRA  & 3.28 & [2.53, 4.15] & 8/8 \\
\method\ -- SAPT      & 3.17 & [0.61, 6.24] & 6/8 \\
\bottomrule
\end{tabular}
\caption{Paired task-level bootstrap over final task scores. Differences use seed-averaged task scores. Confidence intervals use 10,000 resamples of the 8 primary-order tasks.}
\label{tab:paired-bootstrap}
\end{table}

\section{Acquisition and Isolated References}
\label{app:reference-details}

Tables~\ref{tab:appendix-framework-comparison} and \ref{tab:appendix-backbone-comparison} pair sequential acquisition (SEQ) with isolated reference performance (ISO). Each table has 2 panels of 4 tasks. Shaded columns report the signed difference $\Delta=\mathrm{SEQ}-\mathrm{ISO}$. Each difference is calculated from the displayed scores. These columns describe acquisition proximity; final retention is reported separately.

\paragraph{Reference specification.}
The framework comparison uses each method's isolated variant. The backbone comparison uses the reported single-task references; its Qwen3 row matches the \method\ framework reference. The original experimental record describes the backbone references as isolated QLoRA adaptation. Reference adaptation can therefore differ from the sequential method. We treat these gaps as descriptive. Qwen3-8B-R abbreviates Qwen3-8B-Reasoning.

\paragraph{Scope across backbones.}
Table~\ref{tab:appendix-backbone-comparison} gives per-task acquisition and isolated references for all 6 backbones. Table~\ref{tab:backbones} gives final macro scores, forgetting, backward transfer, and forgetting variance. These final-stage measurements assess retention after the full stream.

\begin{table*}[t]
\centering
\small
\setlength{\tabcolsep}{3.7pt}
\renewcommand{\arraystretch}{1.08}
\caption{Acquisition and isolated references for the framework comparison on Qwen3-8B-Reasoning. Scores are means over seeds 37, 73, and 101. Each task group contains SEQ, ISO, and the signed difference in a shaded column.}
\label{tab:appendix-framework-comparison}
\begin{tabular}{l*{4}{rr>{\columncolor{tiladelta}}r}}
\toprule
Method & \multicolumn{3}{c}{FOMC} & \multicolumn{3}{c}{C-STANCE} & \multicolumn{3}{c}{MeetingBank} & \multicolumn{3}{c}{ScienceQA} \\
\cmidrule(lr){2-4}\cmidrule(lr){5-7}\cmidrule(lr){8-10}\cmidrule(lr){11-13}
 & SEQ & ISO & $\Delta$ & SEQ & ISO & $\Delta$ & SEQ & ISO & $\Delta$ & SEQ & ISO & $\Delta$ \\
\midrule
Seq-train & 74.91 & 74.91 & 0.00 & 74.30 & 77.89 & -3.59 & 56.91 & 60.47 & -3.56 & 50.42 & 58.89 & -8.47 \\
L2P & 73.50 & 73.50 & 0.00 & 71.41 & 74.65 & -3.24 & 43.78 & 46.09 & -2.31 & 48.58 & 52.57 & -3.99 \\
GEM & 68.03 & 68.03 & 0.00 & 81.38 & 85.15 & -3.77 & 50.07 & 54.75 & -4.68 & 48.59 & 53.66 & -5.07 \\
A-GEM & 69.54 & 69.54 & 0.00 & 83.89 & 86.26 & -2.37 & 51.62 & 55.80 & -4.18 & 50.42 & 54.31 & -3.89 \\
SAPT & 73.60 & 73.60 & 0.00 & 81.20 & 82.86 & -1.66 & 57.13 & 58.30 & -1.17 & 73.12 & 73.86 & -0.74 \\
InfLoRA & 71.18 & 71.18 & 0.00 & 85.98 & 86.98 & -1.00 & 56.00 & 57.73 & -1.73 & 70.69 & 72.14 & -1.45 \\
SwitchCIT & 73.39 & 73.39 & 0.00 & 84.76 & 86.21 & -1.45 & 55.94 & 56.94 & -1.00 & 73.51 & 75.26 & -1.75 \\
QLoRA-CT & 72.87 & 72.87 & 0.00 & 81.48 & 82.33 & -0.85 & 54.37 & 54.95 & -0.58 & 67.82 & 70.04 & -2.22 \\
MoCL      & 73.79 & 73.79 & 0.00 & 83.14 & 83.84 & -0.70 & 54.74 & 55.25 & -0.51 & 68.15 & 69.94 & -1.79 \\
TreeLoRA  & 74.71 & 74.71 & 0.00 & 84.80 & 85.34 & -0.54 & 55.10 & 55.55 & -0.45 & 68.49 & 69.84 & -1.35 \\
GainLoRA & 76.80 & 76.80 & 0.00 & 86.24 & 87.65 & -1.41 & 54.07 & 55.12 & -1.05 & 67.51 & 69.75 & -2.24 \\
GoD-MoE   & 75.62 & 75.62 & 0.00 & 86.46 & 86.85 & -0.39 & 55.47 & 55.86 & -0.39 & 68.82 & 69.73 & -0.91 \\
TILA-CT & 76.48 & 76.48 & 0.00 & 88.29 & 88.51 & -0.22 & 55.87 & 56.19 & -0.32 & 69.19 & 69.62 & -0.43 \\
\bottomrule
\end{tabular}
\vspace{0.8em}
\begin{tabular}{l*{4}{rr>{\columncolor{tiladelta}}r}}
\toprule
Method & \multicolumn{3}{c}{NumGLUE-en} & \multicolumn{3}{c}{JEC-QA} & \multicolumn{3}{c}{MedMCQA} & \multicolumn{3}{c}{20Minuten} \\
\cmidrule(lr){2-4}\cmidrule(lr){5-7}\cmidrule(lr){8-10}\cmidrule(lr){11-13}
 & SEQ & ISO & $\Delta$ & SEQ & ISO & $\Delta$ & SEQ & ISO & $\Delta$ & SEQ & ISO & $\Delta$ \\
\midrule
Seq-train & 40.64 & 49.20 & -8.56 & 25.46 & 34.43 & -8.97 & 48.34 & 59.14 & -10.80 & 39.98 & 47.96 & -7.98 \\
L2P & 42.13 & 47.04 & -4.91 & 28.83 & 34.94 & -6.11 & 46.45 & 51.77 & -5.32 & 48.74 & 53.73 & -4.99 \\
GEM & 22.73 & 26.95 & -4.22 & 32.30 & 37.19 & -4.89 & 51.54 & 58.14 & -6.60 & 36.35 & 43.62 & -7.27 \\
A-GEM & 23.57 & 27.86 & -4.29 & 33.14 & 38.37 & -5.23 & 53.81 & 58.90 & -5.09 & 38.85 & 43.71 & -4.86 \\
SAPT & 58.26 & 58.85 & -0.59 & 49.85 & 50.35 & -0.50 & 63.26 & 64.55 & -1.29 & 57.10 & 59.48 & -2.38 \\
InfLoRA & 56.05 & 58.21 & -2.16 & 46.50 & 48.97 & -2.47 & 61.73 & 63.64 & -1.91 & 57.44 & 59.03 & -1.59 \\
SwitchCIT & 56.03 & 58.19 & -2.16 & 48.28 & 49.77 & -1.49 & 63.46 & 64.46 & -1.00 & 57.53 & 58.11 & -0.58 \\
QLoRA-CT & 52.64 & 54.79 & -2.15 & 41.98 & 45.48 & -3.50 & 55.12 & 59.37 & -4.25 & 49.15 & 53.27 & -4.12 \\
MoCL & 53.56 & 55.35 & -1.79 & 43.68 & 46.62 & -2.94 & 57.29 & 60.83 & -3.54 & 52.34 & 55.68 & -3.34 \\
TreeLoRA & 54.48 & 55.90 & -1.42 & 45.39 & 47.75 & -2.36 & 59.47 & 62.28 & -2.81 & 55.52 & 58.09 & -2.57 \\
GainLoRA & 53.81 & 56.26 & -2.45 & 48.12 & 48.16 & -0.04 & 62.74 & 62.94 & -0.20 & 59.33 & 61.43 & -2.10 \\
GoD-MoE & 55.40 & 56.46 & -1.06 & 47.09 & 48.89 & -1.80 & 61.64 & 63.74 & -2.10 & 58.71 & 60.50 & -1.79 \\
TILA-CT & 56.41 & 57.07 & -0.66 & 48.97 & 50.14 & -1.17 & 64.03 & 65.34 & -1.31 & 62.21 & 63.15 & -0.94 \\
\bottomrule
\end{tabular}
\end{table*}

\begin{table*}[t]
\centering
\small
\setlength{\tabcolsep}{3.7pt}
\renewcommand{\arraystretch}{1.08}
\caption{Acquisition and isolated references for \method\ across backbones. Each task group contains SEQ, ISO, and the signed difference in a shaded column. These acquisition gaps are distinct from final retention.}
\label{tab:appendix-backbone-comparison}
\begin{tabular}{l*{4}{rr>{\columncolor{tiladelta}}r}}
\toprule
Backbone & \multicolumn{3}{c}{FOMC} & \multicolumn{3}{c}{C-STANCE} & \multicolumn{3}{c}{MeetingBank} & \multicolumn{3}{c}{ScienceQA} \\
\cmidrule(lr){2-4}\cmidrule(lr){5-7}\cmidrule(lr){8-10}\cmidrule(lr){11-13}
 & SEQ & ISO & $\Delta$ & SEQ & ISO & $\Delta$ & SEQ & ISO & $\Delta$ & SEQ & ISO & $\Delta$ \\
\midrule
Llama-3.2-1B & 69.57 & 69.57 & 0.00 & 69.61 & 70.20 & -0.59 & 62.49 & 64.00 & -1.51 & 63.00 & 63.81 & -0.81 \\
Qwen-1.8B & 67.04 & 67.04 & 0.00 & 69.08 & 69.69 & -0.61 & 61.71 & 62.45 & -0.74 & 63.02 & 63.90 & -0.88 \\
Llama-3.1-8B & 71.72 & 71.72 & 0.00 & 71.76 & 72.37 & -0.61 & 64.42 & 65.98 & -1.56 & 64.95 & 65.79 & -0.84 \\
OLMoE-7B & 73.61 & 73.61 & 0.00 & 73.63 & 75.29 & -1.66 & 69.79 & 72.30 & -2.51 & 65.90 & 67.08 & -1.18 \\
Qwen3-8B-R & 76.48 & 76.48 & 0.00 & 88.29 & 88.51 & -0.22 & 55.87 & 56.19 & -0.32 & 69.19 & 69.62 & -0.43 \\
Falcon-H1R-7B & 83.75 & 83.75 & 0.00 & 91.59 & 92.25 & -0.66 & 75.25 & 76.06 & -0.81 & 79.56 & 80.33 & -0.77 \\
\bottomrule
\end{tabular}
\vspace{0.8em}
\begin{tabular}{l*{4}{rr>{\columncolor{tiladelta}}r}}
\toprule
Backbone & \multicolumn{3}{c}{NumGLUE-en} & \multicolumn{3}{c}{JEC-QA} & \multicolumn{3}{c}{MedMCQA} & \multicolumn{3}{c}{20Minuten} \\
\cmidrule(lr){2-4}\cmidrule(lr){5-7}\cmidrule(lr){8-10}\cmidrule(lr){11-13}
 & SEQ & ISO & $\Delta$ & SEQ & ISO & $\Delta$ & SEQ & ISO & $\Delta$ & SEQ & ISO & $\Delta$ \\
\midrule
Llama-3.2-1B & 66.63 & 67.58 & -0.95 & 68.31 & 69.51 & -1.20 & 68.88 & 70.05 & -1.17 & 68.07 & 69.44 & -1.37 \\
Qwen-1.8B & 65.03 & 66.55 & -1.52 & 66.72 & 67.59 & -0.87 & 66.37 & 68.29 & -1.92 & 66.38 & 67.49 & -1.11 \\
Llama-3.1-8B & 68.70 & 69.67 & -0.97 & 70.43 & 71.66 & -1.23 & 71.00 & 72.21 & -1.21 & 70.17 & 71.59 & -1.42 \\
OLMoE-7B & 67.91 & 70.84 & -2.93 & 66.99 & 69.12 & -2.13 & 71.98 & 73.39 & -1.41 & 71.16 & 72.33 & -1.17 \\
Qwen3-8B-R & 56.41 & 57.07 & -0.66 & 48.97 & 50.14 & -1.17 & 64.03 & 65.34 & -1.31 & 62.21 & 63.15 & -0.94 \\
Falcon-H1R-7B & 71.57 & 72.38 & -0.81 & 74.22 & 75.12 & -0.90 & 73.94 & 74.84 & -0.90 & 73.14 & 74.03 & -0.89 \\
\bottomrule
\end{tabular}
\end{table*}

\subsection{Seed variation}
\label{app:seed-variation}

Gray subscripts report standard deviations across seeds 37, 73, and 101. Table~\ref{tab:mainresults} gives acquisition and final macro standard deviations for all methods. Table~\ref{tab:appendix-final-task-scores} gives final-score standard deviations for the first 7 tasks. Comparisons use mean differences and standard deviations.

\section{Aggregate Routing Diagnostics}
\label{app:router-details}

\begin{figure}[t]
\centering
\includegraphics[width=0.85\linewidth]{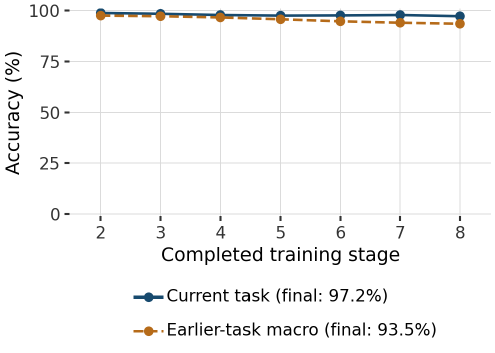}
\caption{Routing accuracy during continual training on Qwen3-8B-Reasoning under the primary task order. Curves show current-task accuracy and past-task macro accuracy. Stage 1 is omitted because only one module is available.}
\label{fig:router}
\end{figure}

Routing accuracy is the percentage of inputs assigned to the correct task module. Table~\ref{tab:appendix-router-summary} reports per-task and aggregate routing accuracy across training stages on the primary CT-Pipeline order. Figure~\ref{fig:router} plots current-task accuracy and past-task macro accuracy. Stage 1 has only one candidate module, so its routing accuracy is 100\% by construction.

\begin{table*}[t]
\centering
\small
\setlength{\tabcolsep}{3.2pt}
\renewcommand{\arraystretch}{1.1}
\resizebox{\textwidth}{!}{%
\begin{tabular}{lcccccccccccc}
\toprule
& \multicolumn{8}{c}{\textbf{Per-task top-1 routing accuracy (\%)}} &
\multicolumn{4}{c}{\textbf{Stage summary (\%)}} \\
\cmidrule(lr){2-9}
\cmidrule(lr){10-13}
Stage
& $T_1$ & $T_2$ & $T_3$ & $T_4$
& $T_5$ & $T_6$ & $T_7$ & $T_8$
& Current & Past macro & All macro & Worst task \\
\midrule
After $T_1$
& 100.0 & -- & -- & -- & -- & -- & -- & --
& 100.0 & -- & 100.0 & 100.0 \\
After $T_2$
& 97.5 & 98.8 & -- & -- & -- & -- & -- & --
& 98.8 & 97.5 & 98.2 & 97.5 \\
After $T_3$
& 96.3 & 98.0 & 98.4 & -- & -- & -- & -- & --
& 98.4 & 97.2 & 97.6 & 96.3 \\
After $T_4$
& 95.5 & 97.6 & 96.6 & 97.8 & -- & -- & -- & --
& 97.8 & 96.6 & 96.9 & 95.5 \\
After $T_5$
& 94.8 & 97.2 & 95.8 & 94.8 & 97.5 & -- & -- & --
& 97.5 & 95.7 & 96.0 & 94.8 \\
After $T_6$
& 94.0 & 96.8 & 95.0 & 93.0 & 94.8 & 97.6 & -- & --
& 97.6 & 94.7 & 95.2 & 93.0 \\
After $T_7$
& 93.2 & 96.5 & 94.3 & 91.5 & 93.0 & 95.5 & 97.8 & --
& 97.8 & 94.0 & 94.5 & 91.5 \\
After $T_8$
& 92.5 & 95.8 & 92.5 & 90.5 & 92.0 & 94.5 & 96.5 & 97.2
& 97.2 & 93.5 & 93.9 & 90.5 \\
\midrule
Routing forgetting
& 7.5 & 3.0 & 5.9 & 7.3 & 5.5 & 3.1 & 1.3 & --
& \multicolumn{4}{l}{
$\overline{F}^{\mathrm{route}}=4.8$
(adjusted: 4.4)
} \\
\bottomrule
\end{tabular}%
}
\caption{Stage-wise top-1 routing accuracy on the primary CT-Pipeline order. Tasks $T_1$--$T_8$ correspond to FOMC, C-STANCE, MeetingBank, ScienceQA, NumGLUE-en, JEC-QA, MedMCQA, and 20Minuten. Past macro excludes the current task. The last row gives acquisition-to-final routing declines.}
\label{tab:appendix-router-summary}
\end{table*}

\paragraph{Routing forgetting.}
The average acquisition-to-final routing decline over tasks 1--7 is 4.8 percentage points. Task 1 has only one candidate at acquisition. Using its first competitive routing accuracy of 97.5\% gives a singleton-adjusted average of 4.4 points. These values are separate from downstream task-score forgetting.

\paragraph{Oracle comparison.}
Oracle routing supplies the correct task module to the trained A0 checkpoints without retraining. Final macro performance increases from 63.98 to 65.19, recovering 1.21 points. This gap measures the downstream cost of learned module selection. Routing accuracy and task-score loss have different scales because an incorrectly selected module can still produce a correct output.

\section{Ablation Implementation and Analysis}
\label{app:ablation-details}

The 8 interventions reported in Table~\ref{tab:appendix-ablation} test routing, adaptation, gathering, residual scaling, and task isolation. \textbf{A0 (full TILA-CT)} uses the complete architecture. \textbf{A1 (oracle routing)} uses the trained A0 checkpoints without retraining and forces each example from task $T_j$ through $Z_j$ and $\{A_{j,\ell}\}$. This measures the performance loss from learned module selection after task paths are frozen.

\textbf{A2 (independent routing keys)} replaces the memory-derived key $k_i=W_k\Mean(Z_i)$ with an independent learned key $k_i=e_i$, where $e_i\in\mathbb{R}^{d_r}$. The remaining architecture is unchanged. We compare macro routing accuracy, worst-task routing accuracy, routing forgetting, and the oracle-routing gap to assess memory-derived prototypes.

\textbf{A3 (standard adapters)} replaces latent gather-and-inject adaptation with private bottleneck MLP adapters. It keeps the input-only sequence router and uses independent keys because latent memories are absent. The bottleneck width is 1536, giving 89.17M parameters for the standard adapter and independent-key router, compared with 90.31M for full \method\ including its router. Comparing A2 with A3 keeps the key design fixed and tests latent adaptation against a private residual MLP. QLoRA-CT uses a different adapter architecture and reference-output loss for selection, so it does not give this control.

\textbf{A4 (inject only)} removes gathering and injects the static learned memory $Z_i$ directly. This tests whether contextualizing memory with each token's causal prefix improves over static memory injection. \textbf{A5 (fixed residual coefficient)} sets every $\gamma_{i,\ell}=1$. Full TILA-CT initializes these coefficients at 0.1 to limit the initial residual and learns them during task training.

\textbf{A6 (shared adapter bank)} keeps private memories and shares the adapter bank across tasks. Each stage updates the current memory, router, and shared bank while earlier memories stay frozen. Later updates therefore change the adapter computation used by earlier tasks. This tests adapter isolation. \textbf{A7 (single shared module)} continually updates one shared memory and adapter bank. It uses no router and removes task isolation. It also uses fewer cumulative parameters, so the comparison tests isolation and capacity together.

\paragraph{Arithmetic convention.}
Scores and derived summaries are rounded to 2 decimal places. Applying the identities to rounded aggregate values can introduce small differences. Oracle gaps compare learned and oracle routing at final evaluation; acquisition and final macro scores are separate measurements. The A3 and A5 oracle gaps are 1.80 and 1.20 points.

\paragraph{Component comparisons.}
A0 versus A2 tests memory-derived prototypes. The 0.18-point acquisition difference grows to 0.66 at final evaluation. A0 has 3.4 percentage points higher macro routing accuracy, 5.0 points higher worst-task routing accuracy, and 2.1 points less singleton-adjusted routing forgetting. Its oracle gap is 0.48 points smaller. These diagnostics connect the memory-derived key with stronger module access.

A2 versus A3 tests latent adaptation with independent keys held fixed. The latent adapter gains 1.48 acquisition points and 1.60 final points over the approximately matched MLP. A0 versus A4 tests causal-prefix gathering against static-memory injection. Gathering gains 1.08 acquisition points and 1.19 final points, with 0.13 less forgetting. These comparisons show higher task adaptation scores.

A0 versus A5 tests learned residual scaling. Learning the coefficient gains 0.62 acquisition points and 0.60 final points. Forgetting is 1.38 for A0 and 1.36 for A5. The endpoints show a task-performance gain; training trajectories would be needed to test the initialization-stability motivation.

A6 and A7 test shared computation. A6 keeps private frozen memories, but later tasks update the adapter bank used with them. Adding its reported oracle gap to final performance gives 61.19, placing final oracle performance 3.27 points below its reported acquisition score. A7 has no module-selection step. Its acquisition-to-final decline is 6.48 points, corresponding to 7.40 forgetting over the first 7 tasks. These variants change the preserved computation used by earlier tasks. Their lower cumulative capacity also affects the comparison, so the complete deficits cannot be assigned uniquely to interference.

\onecolumn

\section{Dataset Configurations}
\label{app:data}

\begin{table}[H]
\centering
\setlength{\tabcolsep}{3.2pt}
\renewcommand{\arraystretch}{1}
\caption{Dataset configurations, official splits, output formats, answer extraction rules, and primary metrics used in CT-Pipeline.}
\label{tab:dataset_metrics}
\resizebox{\textwidth}{!}{%
\begin{tabular}{
>{\raggedright\arraybackslash}p{1.8cm}
>{\raggedright\arraybackslash}p{3.0cm}
>{\raggedright\arraybackslash}p{3.2cm}
>{\raggedright\arraybackslash}p{3.0cm}
>{\raggedright\arraybackslash}p{6.1cm}
>{\raggedright\arraybackslash}p{3.2cm}
}
\toprule
\textbf{Dataset}
&
\textbf{Splits used}
&
\textbf{Task type}
&
\textbf{Output format}
&
\textbf{Answer extraction}
&
\textbf{Evaluation metric}
\\
\midrule

FOMC
&
Train: 1,984;\newline
Test: 496
&
English monetary-policy stance classification with 3 classes
&
One canonical label:
\texttt{HAWKISH},
\texttt{DOVISH}, or
\texttt{NEUTRAL}
&
Leading and trailing whitespace and formatting tokens are removed. The output is converted to uppercase and accepted only when exactly one canonical label is present. Source labels are mapped as \texttt{0 = DOVISH}, \texttt{1 = HAWKISH}, and \texttt{2 = NEUTRAL}.
&
Accuracy.
\\

\addlinespace

C-STANCE
&
Train: 33,418;\newline
Validation: 7,284;\newline
Test: 7,424
&
Chinese target-conditioned stance classification with 3 classes
&
One canonical label:
\texttt{FAVOR},
\texttt{AGAINST}, or
\texttt{NEUTRAL}
&
Formatting tokens are removed and the final canonical stance label is extracted. Equivalent forms may be normalized, such as \texttt{PRO} to \texttt{FAVOR}, \texttt{CON} to \texttt{AGAINST}, and \texttt{NEU} to \texttt{NEUTRAL}. Free-form reasoning text is ignored during scoring.
&
Accuracy.
\\

\addlinespace

Meeting-Bank
&
Train: 5,169;\newline
Validation: 861;\newline
Test: 862
&
English meeting-segment abstractive summarization
&
Free-form English summary
&
Chat-template artifacts, unintended answer prefixes, and leading or trailing whitespace are removed. The generated summary content and sentence order are otherwise preserved.
&
ROUGE-L F1 on a 0--100 scale.
\\

\addlinespace

ScienceQA
&
Train: 12,726;\newline
Validation: 4,241;\newline
Test: 4,241
&
Multiple-choice science question answering using the text-only subset
&
One option identifier, usually
\texttt{A} through \texttt{E}
&
The response following the final-answer marker is extracted. Option letters and valid numerical option indices are mapped to the corresponding answer choice. Only one final option is accepted. Image-dependent examples are excluded in the text-only setting, and the post-filter split counts are reported separately.
&
Answer-option accuracy.
\\

\addlinespace

NumGLUE-en
&
Train: 5,000;\newline
Test: 2,000;\newline
Official splits used by TRACE \citep{wang2023trace}
&
English numerical and arithmetic reasoning
&
A short numerical answer, such as an integer, decimal, signed number, or percentage
&
The response following the final-answer marker is extracted. Whitespace, commas, signs, decimal forms, percentage symbols, units, and trailing punctuation are normalized by the selected evaluator. Intermediate calculations are ignored.
&
Normalized exact-match accuracy. For categorical NumGLUE subtasks, the official label accuracy is used.
\\

\addlinespace

JEC-QA
&
Train: 21,902;\newline
Test: 5,273
&
Chinese legal-domain multiple-choice question answering with single-answer and multiple-answer questions
&
A canonical option set, such as
\texttt{A},
\texttt{C},
\texttt{AB}, or
\texttt{BCD}
&
The response is converted to uppercase and only valid option identifiers are retained. Duplicate options are removed, and the predicted set is sorted into canonical order. A prediction is correct only when the full predicted option set matches the gold option set exactly. Missing or extra choices are marked incorrect.
&
Exact option-set accuracy.
\\

\addlinespace

MedMCQA
&
Train: 182,822;\newline
Test: 6,150;\newline
Official validation set used as test set
&
English medical multiple-choice question answering with 4 options
&
One option identifier:
\texttt{A},
\texttt{B},
\texttt{C}, or
\texttt{D};
equivalently, an index from 1 to 4
&
Formatting tokens are removed and the final option is mapped to the official one-based answer index. Outputs containing multiple distinct answer choices are marked incorrect.
&
Answer-option accuracy.
\\

\addlinespace

20Minuten
&
Train: 17,905;\newline
Development: 200;\newline
Test: 200
&
German document-level text simplification and content compression
&
Free-form simplified German text
&
Chat wrappers, unintended response prefixes, and leading or trailing whitespace are removed. Lexical, syntactic, and content changes made by the model are preserved for evaluation.
&
SARI.
\\

\bottomrule
\end{tabular}%
}
\end{table}

\clearpage

\section{Additional Result Visualizations}

\begin{figure}[H]
  \centering
  \begin{minipage}[b]{0.49\textwidth}
    \centering
    \includegraphics[width=\linewidth]{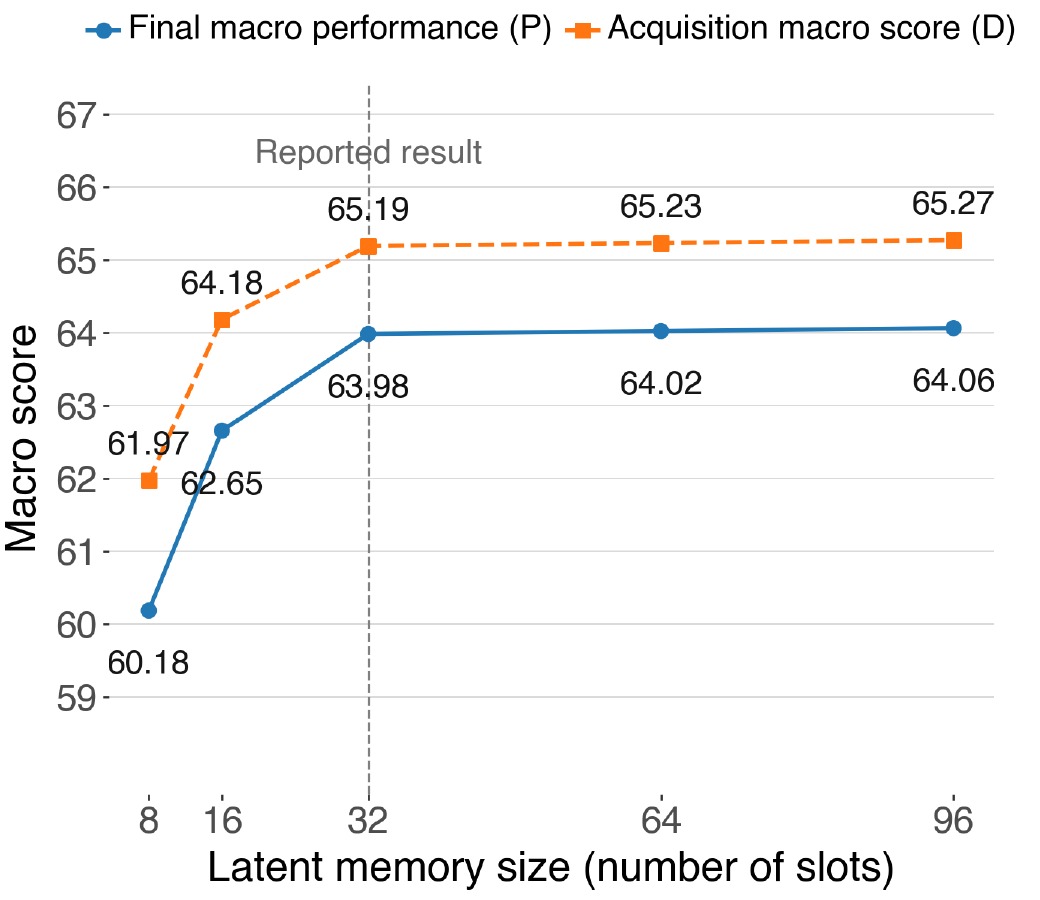}
    \par\smallskip
    {\small (a) Memory: macro scores}
  \end{minipage}\hfill
  \begin{minipage}[b]{0.42\textwidth}
    \centering
    \includegraphics[width=\linewidth]{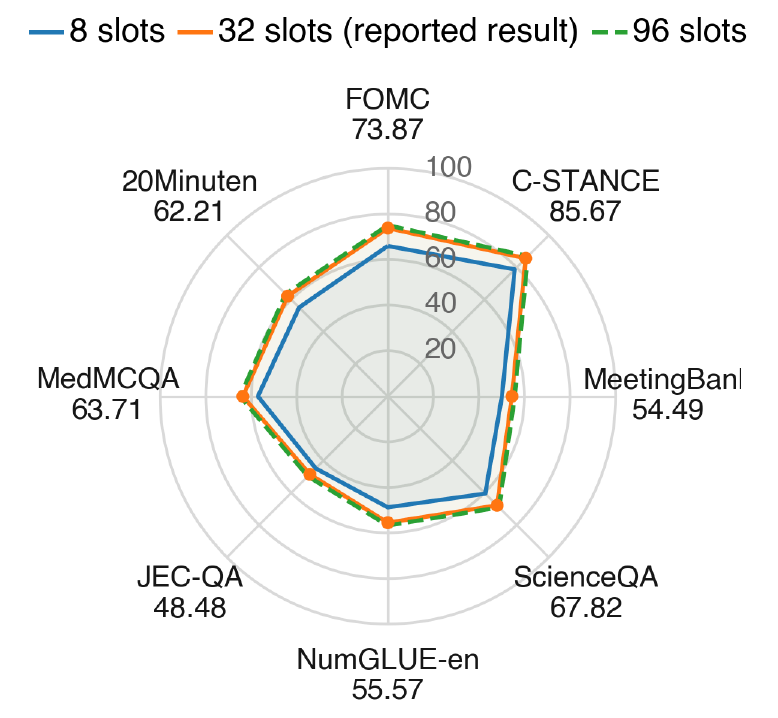}
    \par\smallskip
    {\small (b) Memory: task scores}
  \end{minipage}

\caption{Latent-memory capacity sensitivity. The line plot shows acquisition $D$ and final macro performance $P$ as the number of memory slots $L_z$ changes. The radar plot shows final task scores for selected memory sizes.}
  \label{fig:memory-sensitivity}
\end{figure}

\begin{figure}[H]
  \centering
  \begin{minipage}[b]{0.49\textwidth}
    \centering
    \includegraphics[width=\linewidth]{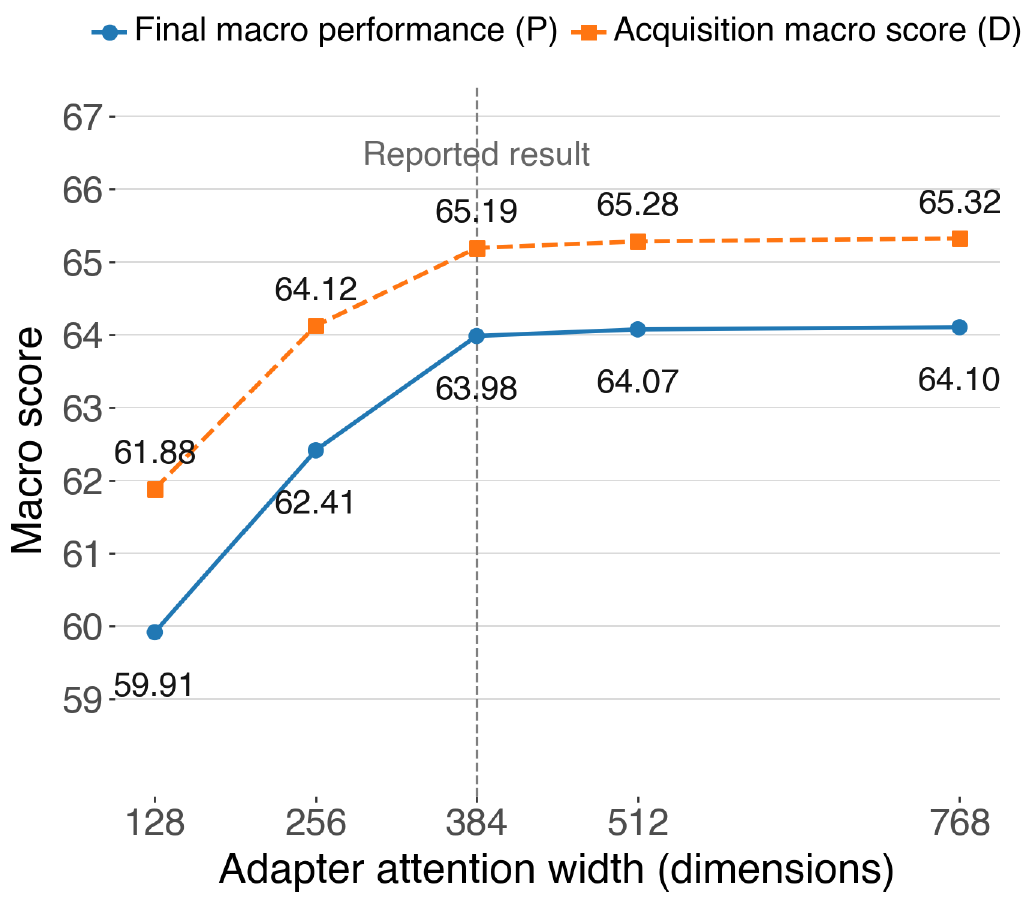}
    \par\smallskip
    {\small (a) Width: macro scores}
  \end{minipage}\hfill
  \begin{minipage}[b]{0.42\textwidth}
    \centering
    \includegraphics[width=\linewidth]{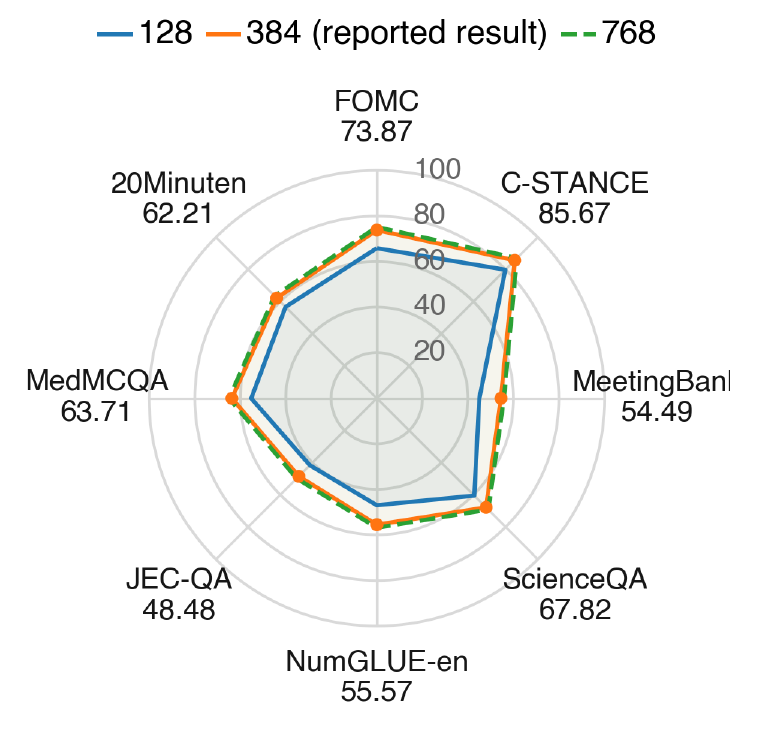}
    \par\smallskip
    {\small (b) Width: task scores}
  \end{minipage}

\caption{Adapter-width capacity sensitivity. The line plot shows acquisition $D$ and final macro performance $P$ as adapter attention width $d_a$ changes. The radar plot shows final task scores for selected widths.}
  \label{fig:width-sensitivity}
\end{figure}

\begin{figure}[H]
  \centering
  \includegraphics[width=0.90\linewidth]{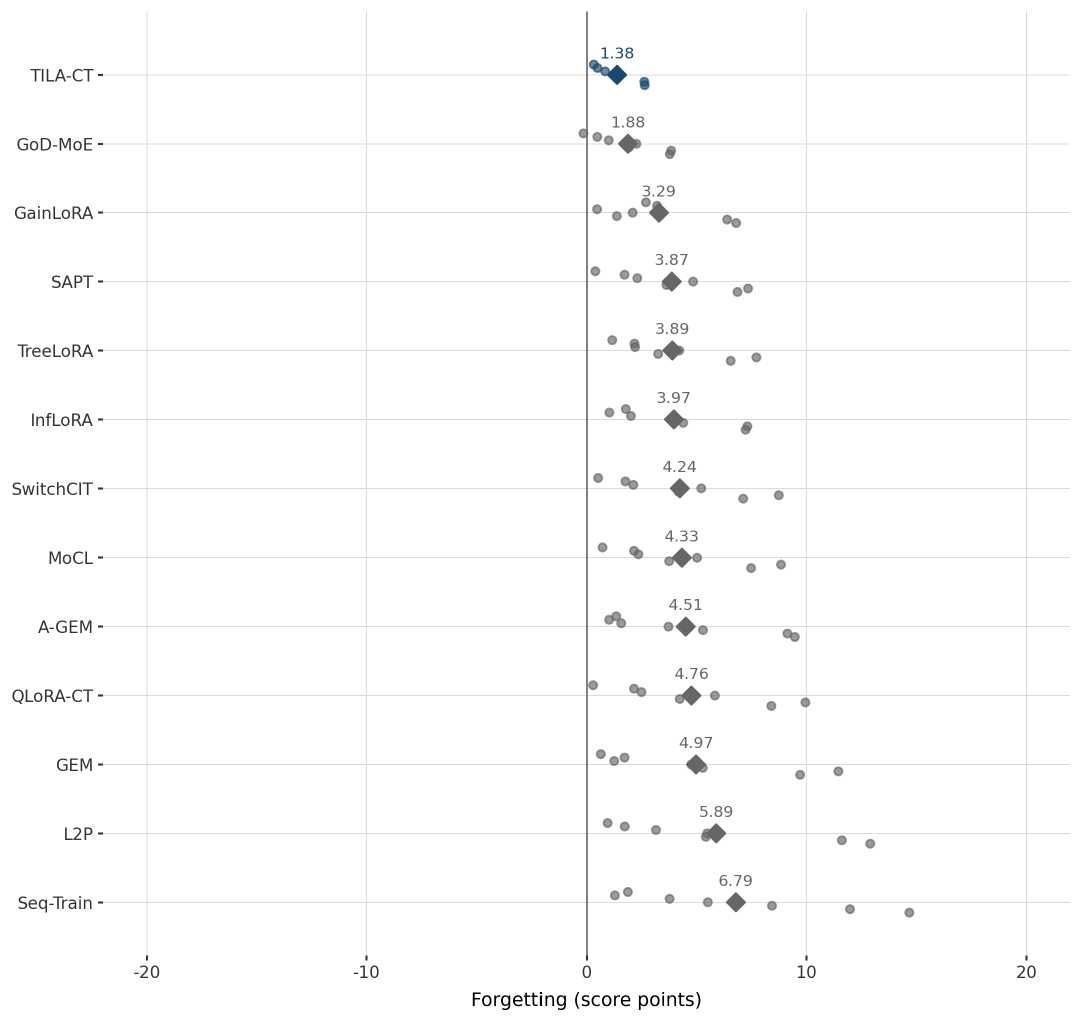}
  \caption{Forgetting across the first 7 tasks on Qwen3-8B-Reasoning. Each dot is one task's acquisition score minus its final score, computed from scores averaged over seeds 37, 73, and 101. Diamonds mark the 7-task mean. Negative values indicate improvement after acquisition. The final task is excluded because it has no subsequent training stage. The x-axis spans $-20$ to $20$ points.}
  \label{fig:forgetting}
\end{figure}

\end{document}